\documentclass[10pt,twocolumn,letterpaper]{article}

\usepackage[pagenumbers]{cvpr} 

\makeatletter
\@namedef{ver@everyshi.sty}{}
\makeatother
\usepackage{graphicx}
\usepackage{amsmath}
\usepackage{amssymb}
\usepackage{booktabs}
\usepackage{microtype}
\usepackage{pifont}
\usepackage{color}
\usepackage{placeins}
\usepackage{float}
\usepackage{soul}
\usepackage{enumitem}
\usepackage{adjustbox}
\usepackage{commath}
\usepackage{bbm}
\usepackage{multirow}
\usepackage{caption}
\usepackage{longtable}
\usepackage{array}
\usepackage{xcolor}
\usepackage{makecell}
\usepackage{subcaption}
\usepackage{tcolorbox}
\usepackage[accsupp]{axessibility}

\newcommand{\cmark}{\ding{51}}%
\newcommand{\xmark}{\ding{55}}%

\usepackage[pagebackref,breaklinks,colorlinks]{hyperref}

\def\confName{CVPR}
\def\confYear{2023}

\begin{document}

\title{Language in a Bottle: Language Model Guided Concept Bottlenecks \\ for
Interpretable Image Classification}

\author{Yue Yang, Artemis Panagopoulou, Shenghao Zhou, Daniel Jin, \\Chris Callison-Burch, Mark Yatskar\\
University of Pennsylvania\\
{\tt\small \{yueyang1, artemisp, shzhou2, jindan, ccb, myatskar\}@seas.upenn.edu}
}

\maketitle



\begin{abstract}

Concept Bottleneck Models (CBM) are inherently interpretable models that factor model decisions into human-readable concepts.
They allow people to easily understand why a model is failing, a critical feature for high-stakes applications.
CBMs require manually specified concepts and often under-perform their black box counterparts, preventing their broad adoption.
We address these shortcomings and are first to show how to construct high-performance CBMs without manual specification of similar accuracy to black box models.
Our approach, \textbf{La}nguage Guided \textbf{Bo}ttlenecks (LaBo), leverages a language model, GPT-3, to define a large space of possible bottlenecks.
Given a problem domain, LaBo uses GPT-3 to produce factual sentences about categories to form candidate concepts.
LaBo efficiently searches possible bottlenecks through a novel submodular utility that promotes the selection of discriminative and diverse information.
Ultimately, GPT-3's sentential concepts can be aligned to images using CLIP, to form a bottleneck layer.
Experiments demonstrate that LaBo is a highly effective prior for concepts important to visual recognition. 
In the evaluation with 11 diverse datasets, LaBo bottlenecks excel at few-shot classification: they are 11.7\% more accurate than black box linear probes at 1 shot and comparable with more data.
Overall, LaBo demonstrates that inherently interpretable models can be widely applied at similar, or better, performance than black box approaches.\footnote{Code and data are available at \href{https://github.com/YueYANG1996/LaBo}{https://github.com/YueYANG1996/LaBo}}




\end{abstract}

\section{Introduction}


As deep learning systems improve, their applicability to critical domains is hampered because of a lack of transparency.
Efforts to address this have largely focused on post-hoc explanations~\cite{selvaraju2017grad, Lime_ribeiro2016should, zhang2021survey}.
Such explanations can be problematic because they may be incomplete or unfaithful with respect to the model's computations~\cite{rudin2019stop}.
Models can also be designed to be inherently interpretable, but it is believed that such models will perform more poorly than their black box alternatives~\cite{gunning2019darpa}. 
In this work, we provide evidence to the contrary.
We show how to construct high-performance interpretable-by-design classifiers by combining a language model, GPT-3~\cite{NEURIPS2020_1457c0d6}, and a language-vision model, CLIP~\cite{radford2021learning}.


\begin{figure}[!t]
\centering
    \includegraphics[width=8.3cm]{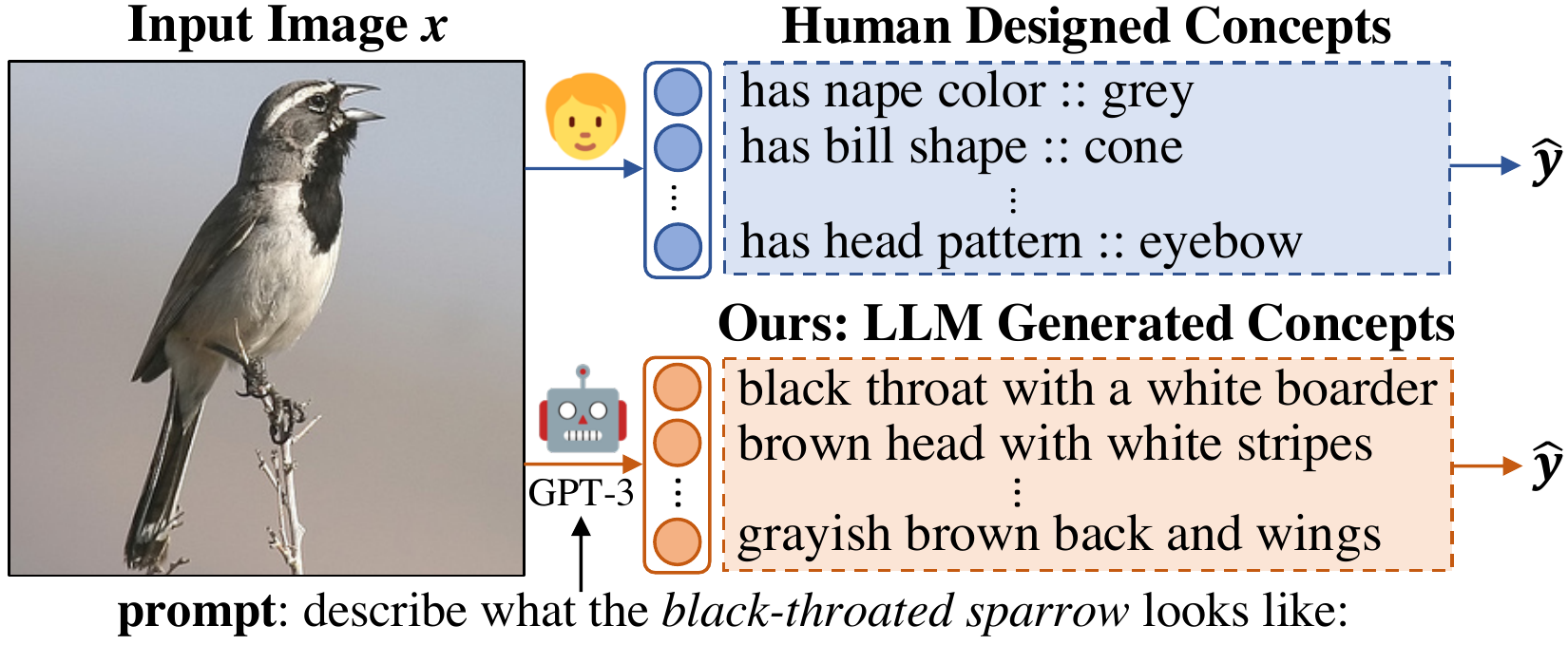}
    \caption{Our proposed high-performance Concept Bottleneck Model alleviates the need for human-designed concepts by prompting large language models (LLMs) such as GPT-3\cite{NEURIPS2020_1457c0d6}.
    }
    \label{fig: intro example}
    \vspace{-0.4cm}
\end{figure}

\begin{figure*}[!t]
    \centering
    \includegraphics[width=16.5cm]{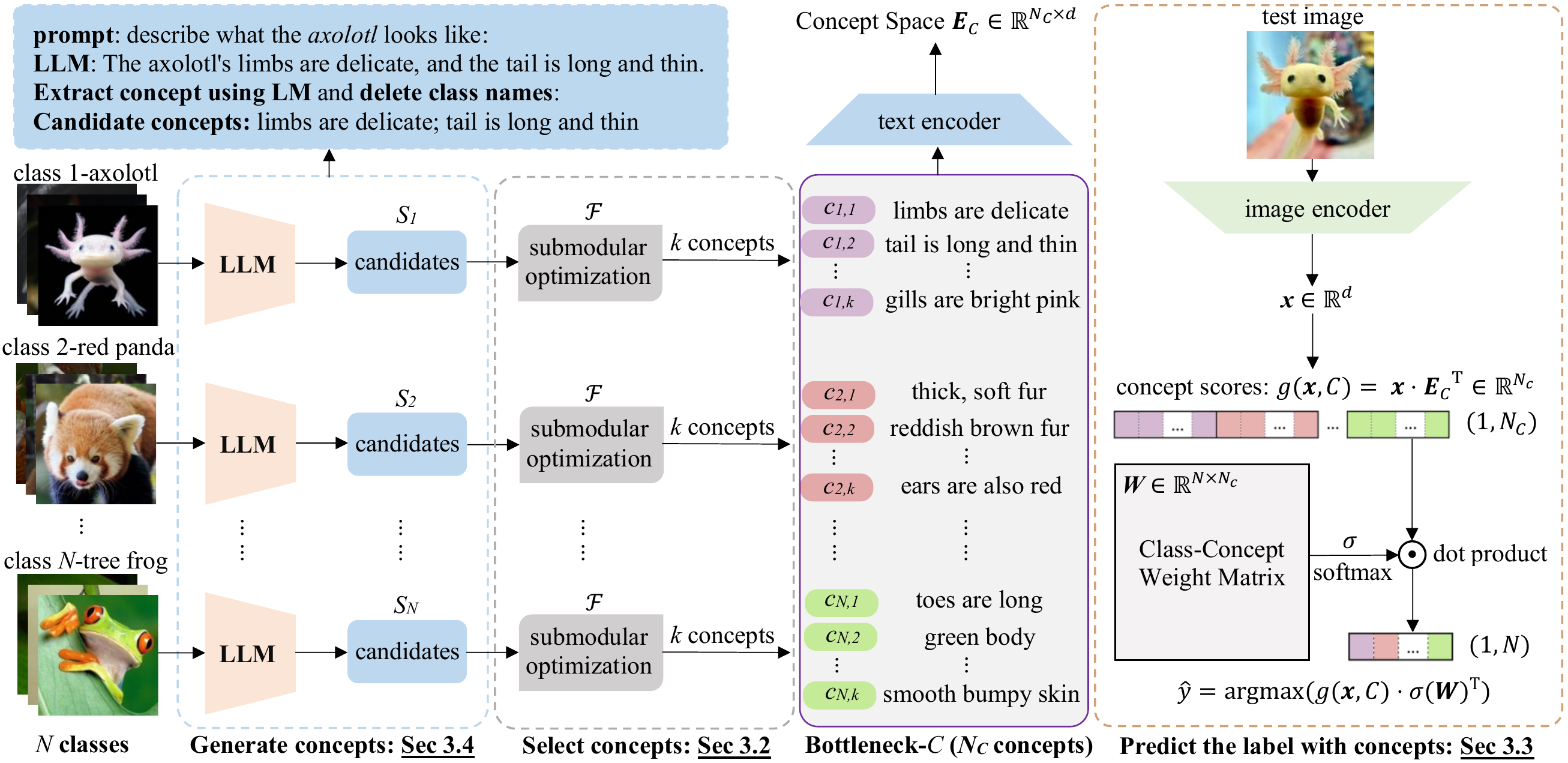}
    \caption{We present an overview of our \textbf{La}nguage-Model-Guided Concept \textbf{Bo}ttleneck Model (\textbf{LaBo}), which is interpretable by design image classification system. First, we prompt the large language model (GPT-3) to generate candidate concepts (Sec \ref{sec: generation}). Second, we employ a submodular function to select concepts from all candidates to construct the bottleneck (Sec \ref{sec: concept selection}). Third, we apply a pretrained alignment model (CLIP) to obtain the embeddings of concepts and images, which is used to compute concept scores. Finally, we train a linear function in which the weight $\boldsymbol{W}$ denotes the concept-class association user to predict targets based on concept scores (Sec \ref{sec: linear function}).}
    \label{fig: diagram}
    \vspace{-0.3cm}
\end{figure*}

Our method builds on Concept Bottleneck Models (CBM)~\cite{koh2020concept}, which construct predictors through a linear combination of human-designed concepts.
For example, as seen in Figure~\ref{fig: intro example}, a qualified person can design concepts, such as ``nape color,'' as intermediate targets for a black box model before classifying a bird.
CBMs provide abstractions that people can use to understand errors or intervene on, contributing to increased trust.

Application of CBMs is limited because they require costly attribute annotations by domain experts and often under-perform their black box counterparts. 
In contexts where CBM performance is competitive with black box alternatives, interpretability properties are sacrificed~\cite{margeloiu2021concept,yuksekgonul2022post}.
To address both of these challenges, we propose to build systems that automatically construct CBMs.  

Our \textbf{La}nguage Model Guided Concept \textbf{Bo}ttleneck Model (\textbf{LaBo}), Figure~\ref{fig: diagram}, allows for the automatic construction of high-performance CBMs for arbitrary classification problems without concept annotations. 
Large language models (LLMs) contain significant world knowledge~\cite{petroni-etal-2019-language,jiang-etal-2020-know,sung-etal-2021-language}, that can be elicited by inputting a string prefix and allowing LLMs to complete the string (prompting). 
For example, in Figure~\ref{fig: intro example}, GPT-3 is prompted about sparrows and completes with information such as ``brown head with white stripes.'' 
LaBo leverages this by constructing bottlenecks where the concepts are such GPT-3 generated sentences. 
Since our concepts are textual, we use CLIP to score their presence in an image and form a bottleneck layer out of these scores.

A key advantage of LaBo is the ability to control the selection of concepts in the bottleneck by generating candidates from the language model.
We develop selection principles targeting both interpretability and classification accuracy.
For example, we prefer smaller bottlenecks that include shorter sentences that do not include class names.
Furthermore, to maximize performance, we prefer attributes that CLIP can easily recognize and are highly discriminative.
To account for appearance variation, we select attributes that cover a variety of information and are not repetitive. 
We formulate these factors into a novel sub-modular criterion that allows us to select good bottlenecks efficiently~\cite{nemhauser1978analysis}. 

We have evaluated LaBo-created bottlenecks on 11 diverse image classification tasks, spanning recognition of common objects ~\cite{deng2009imagenet,krizhevsky2009learning} to skin tumors~\cite{tschandl2018ham10000}.
fine-grained types~\cite{wah2011caltech, Nilsback08,bossard14,maji2013fine}, textures~\cite{cimpoi2014describing}, actions~\cite{soomro2012ucf101}, skin tumors~\cite{tschandl2018ham10000}, and satellite photographed objects~\cite{cheng2017remote}.\footnote{The only dataset specialization we perform is prompt tuning for GPT-3 when creating candidate attributes. This is largely done to overcome problems of word sense. For example, when naively prompted to produce knowledge about the flower ``bird of paradise'' GPT-3 yields information about birds instead of flowers. In general, specialization here was also minimal. See appendix for prompts.}
Our main finding is that LaBo is a highly effective prior for what concepts to look for, especially in low data regimes.
In evaluations comparing with linear probes, LaBo outperforms by as much as 11.7\% at 1-shot and marginally underperforms given larger data settings.
Averaged over many dataset sizes, LaBo bottlenecks are 1.5\% more accurate than linear probes. 
In comparison to modifications of CBMs that improve performance by circumventing the bottleneck~\cite{yuksekgonul2022post}, we achieve similar or better results without breaking the CBM abstraction.
In extensive ablations, we study key trade-offs in bottleneck design and show our selection criteria are crucial and highlight several other critical design choices.

Human evaluations indicate that our bottlenecks are largely understandable, visual, and factual. 
Finally, annotators find our GPT-3 sourced bottlenecks are more factual and groundable than those constructed from WordNet or Wikipedia sentences.
Overall, our experiments demonstrate that automatically designed CBMs can be as effective as black box models while maintaining critical factors contributing to their interpretability.

\section{Related Work}
\noindent

Broadly, interpretability methods fall into two categories: \textit{post-hoc} and \textit{by design}. While ours is an instance of the latter, \textbf{post-hoc methods} have the advantage of not imposing any model constraints. For example, \textit{Gradient-weighted Class Activation Mapping} approaches\cite{selvaraju2017grad,bau2017network,mu2020compositional,hernandez2021natural} trace network gradients to identify the input areas that guide predictions. Similarly, \textit{Explanation Generation} methods\cite{hendricks2016generating,kim2018textual,singh2022explaining,nishida2022improving} require models to produce 
explanations for visual tasks by conditioning their predictions on 
captioning models and \cite{hendricks2018grounding,park2018multimodal} incorporate visual evidence to ground explanations. 

Despite their advantages, there is no guarantee that post-hoc methods faithfully represent model reasoning\cite{rudin2019stop}.  In contrast, our work falls under \textbf{interpretable by design methods}, which constrain explanations to align with the model's reasoning. For example, \textit{Prototype} methods\cite{chen2019looks,nauta2021neural,snell2017prototypical,vinyals2016matching,garcia2018fewshot} optimize a metric space that guides classification by computing distances to prototype representations of each class. While such methods identify important regions in the input for classification, they still require featurized 
region representations that obfuscate the semantic content of the region.


This work extends another family of interpretable by design methods known as \textit{Concept Bottleneck Models}  \cite{koh2020concept,sawada2022concept}. 
Following early attempts in few shot learning\cite{lampert2013attribute} and attribute learning\cite{russakovsky2010attribute,xu2020attribute}, CBMs predict targets by linearly combining an intermediate layer of human-understandable attributes. Recently, Computational Derivation Learning (CompDL) \cite{yun2022vision} proposed a CBM architecture that applies a linear layer over CLIP scores between human expert designed concepts and images to predict targets in the context of an evaluation framework to measure how well CLIP grounds concepts. CBMs generally suffer from the need for costly class description annotations and lower performance compared to  end-to-end counterparts. 
Post-hoc Concept Bottleneck (PCBM)\cite{yuksekgonul2022post} was proposed to fill these two gaps by leveraging information from a static knowledge base, such as ConceptNet\cite{speer2017conceptnet}, and adding a residual connection from image features to the final prediction to improve accuracy \cite{yuksekgonul2022post}.
However, PCBMs cannot be expanded to larger-scale (e.g., ImageNet\cite{deng2009imagenet}) or domain-specific tasks (e.g., fine-grained\cite{maji2013fine}) because knowledge bases have limited coverage. 
In addition, they include a residual predictor, which effectively ensembles CBM with an end-to-end model, 
undermining interpretability.

Inspired by previous work on using textual knowledge to guide vision models\cite{bujwid2021large, kil-chao-2021-revisiting, shen2022k, roth2022integrating}, we circumvent the requirement for external knowledge bases, and instead query LLMs to collect concepts. We remove the need for direct mapping from image features to targets by fully automating the extraction and filtering of LLM knowledge.
Our model surpasses end-to-end models in few shot scenarios and achieves comparable performance in large data settings, while concurrent work \cite{menon2022visual} only evaluates on zero-shot settings.

Our work capitalizes on improvements in \textbf{vision-language pretraining} from  earlier BERT-based models\cite{li2019visualbert,tan2019lxmert,lu2019vilbert,chen2020uniter} to more scalable contrastive architectures\cite{radford2021learning,jia2021scaling,yuan2021florence,li2022blip}, which are very effective for few shot image classification\cite{chowdhury2021few,tian2020rethinking}. 

Our work can be viewed as interpretability-focused prompt tuning of CLIP\cite{radford2021learning}. Significant efforts have been devoted to \textbf{prompting} vision language models\cite{deng2022learning,zhou2022learning,zhou2022conditional,gao2021clip,li2022supporting,rao2022denseclip,pratt2022does}. These focus on searching over text prompts to improve classification performance, and resemble earlier techniques in LLM prompt tuning\cite{shin2020autoprompt,schick2021exploiting,gao2021making}.

\vspace{-.2cm}
\section{Method}
Figure \ref{fig: diagram} presents an overview of our method. Our model prompts a large language model, GPT-3\cite{NEURIPS2020_1457c0d6} to generate a set of candidate concepts for each class (Section \ref{sec: generation}). We employ  submodular optimization  to greedily select a subset of concepts for each class such that we maximize discriminability and diversity (Section \ref{sec: concept selection}). We then align the selected concepts to images using CLIP\cite{radford2021learning}. We apply a linear layer over the similarity scores of concepts and images to learn a weight matrix representing the importance of each concept in the final classification. This weight matrix is initialized using a language model prior from GPT-3 (Section \ref{sec: linear function}). 

\subsection{Problem Formulation}
Consider a training set of image-label pairs $\mathcal{D} = \{(i, y)\}$ where $i$ is the image and $y\in \mathcal{Y}$, is a label from a set of $N$ classes. Suppose we have a pretrained multimodal alignment model (e.g., CLIP \cite{radford2021learning}), which has an image encoder $\mathcal{I}$ and a text encoder $\mathcal{T}$. $\mathcal{I}$ and $\mathcal{T}$ can map images and text into the shared feature space, respectively. The dot product of the image and text features reflects the alignment score between the two modalities. We extract the features of all images in $\mathcal{D}$ as $\boldsymbol{x} = \mathcal{I}(i) \in \mathbb{R}^d$, and the dataset can be represented as $\mathcal{D} = \{(\boldsymbol{x}, y)\}$.
Let $S$ be the superset of candidate textual concepts generated from language models. We use a submodular function $\mathcal{F}$ to select a bottleneck, $C$, where $C \subseteq S$, made of $N_C$ concepts, $C = \{c_1, c_2, ..., c_{N_C}\}$.
We can construct a bottleneck embedding, $\boldsymbol{E}_C \in \mathbb{R}^{N_C \times d}$, and each row of $\boldsymbol{E}_C$ is the text feature $\mathcal{T}(c) \in \mathbb{R}^d$ of a concept $c$ extracted by the text encoder $\mathcal{T}$.

Concept bottleneck models produce a prediction by composing two functions, $\hat{y} = f\left(g\left(\boldsymbol{x}, \boldsymbol{E}_C\right)\right)$, in which $g: \mathbb{R}^d \rightarrow \mathbb{R}^{N_C}$ maps the image feature to a score for every element of the bottleneck and $f: \mathbb{R}^{N_C} \rightarrow \mathcal{Y}$ makes the final prediction on the label space given the concept scores. In our setting, we find a  bottleneck $C$ and appropriate $f$ by solving the following minimization problem:
\begin{equation}
    \min_{f, C} \mathop{\mathbb{E}}_{(\boldsymbol{x}, y) \sim \mathcal{D}} \left[\mathcal{L}\left(f\left(g\left(\boldsymbol{x}, \boldsymbol{E}_C\right)\right), y\right)\right] - \mathcal{F}\left(C, \mathcal{D}\right)
\end{equation}
in which $\mathcal{L}(\hat{y}, y)$ is the cross-entropy loss on the label prediction and $\mathcal{F}(C,\mathcal{D})$ is the quality of the bottleneck as measured by the submodular function. 
In practice, we optimize sequentially: we first find a high scoring $C$ under $\mathcal{F}$. Then, we use the dot product of image and concept embeddings as $g$. Finally, we find an $f$ that minimizes $\mathcal{L}$.
In the following sections, we will illustrate how we: construct the submodular function $\mathcal{F}$ to select a subset of concepts $C$ from the candidates $S$ (Section~\ref{sec: concept selection}) and learn $f$ (Section~\ref{sec: linear function}).

\subsection{Submodular Concept Selection} \label{sec: concept selection}
We create a superset of candidate concepts, $S$, out of class-specific subsets. For every label $y \in \mathcal{Y}$, we construct $S_y$ by prompting a language model to produce textual knowledge about $y$ (Section \ref{sec: generation}).
Instead of directly choosing $N_C$ concepts from $S$, we select $k$ concepts for each class, such that $N \times k = N_C$, to ensure each class has an equal number of relevant concepts in the bottleneck. 

We employ submodular optimization \cite{bach2010convex} to select a subset $C_y \subseteq S_y$, $|C_y| = k$. 
Specifically, we need to design a score function $\mathcal{F}: 2^{|S_y|} \rightarrow \mathbb{R}$ to evaluate the utility of the subset. 
Submodular functions should satisfy the \textit{diminishing returns} property.\footnote{\textit{diminishing returns} property means 
$\forall A \subseteq B \subseteq V \setminus v$, we have $\mathcal{F}(A + \{v\}) - \mathcal{F}(A) \geq \mathcal{F}(B + \{v\}) - \mathcal{F}(B)$.} 
If a submodular function is \textit{monotone},\footnote{A submodular function is \textit{monotone} if $\forall A \subseteq B$, $\mathcal{F}(A) \leq \mathcal{F}(B)$.} a greedy algorithm \cite{nemhauser1978analysis} can be used to find a solution within a constant factor of the optimal one.
We propose the following monotone submodular function\footnote{Any linear combination of submodular functions is still submodular.} to select the subset $C_y$ from the candidate set $S_y$:
\begin{equation} \label{eq: submodular function}
    \mathcal{F}(C_y) = \underbrace{\alpha \cdot \sum_{c \in C_y} D(c)}_\text{discriminability} + \underbrace{\beta \cdot \sum_{c_1 \in S_y} \max_{c_2 \in C_y} \phi (c_1, c_2)}_\text{coverage},
\end{equation}
where $D(c)$ denotes the discriminability score of the concept $c$ and $\phi(\cdot)$ is the intra-concept similarity. 
Generally, the first term tends to select more informative concepts, and the second term ensures the subset has good coverage of the candidate set. 
The hyperparameters $\alpha$ and $\beta$ control the weights of the two sub-functions. Here we present how to compute these two scores:
\medbreak
\noindent \textbf{Discriminability Score.}
We introduce a discriminability score to encourage the selection of concepts that are aligned with many images in class $y$, but few images in other classes.
We first define the similarity score $Sim\left(y, c\right)$ between a class and concept by taking the mean of the dot product between the images and text features:
\begin{equation} \label{eq; similarity score}
    Sim\left(y , c\right) = \frac{1}{|\mathcal{X}_y|}\sum
    _{\boldsymbol{x} \in \mathcal{X}_y} \boldsymbol{x} \cdot \mathcal{T}(c)^{\top},
 \end{equation}
where $\mathcal{X}_y$ is the set of training images labeled with $y$\footnote{In $N$-way-$K$-shot setting, $|\mathcal{X}_y| = K$.} and $\mathcal{T}$ is the text encoder.
We define the normalized class association, which measures the conditional likelihood of aligning featurized images of a class given a concept's textual embedding, ${\overline{Sim}(y|c)} = Sim(y,c) / \sum_{y' \in \mathcal{Y}} Sim(y',c)$, and compute its negative entropy:
\begin{equation}
    D(c) = \sum_{y' \in Y} \overline{Sim}\left(y' | c\right) \cdot \log \left( \overline{Sim}\left(y' | c\right)\right)
\end{equation}
Maximizing $D(c)$ will result in the selection of concepts that have peaked $\overline{Sim}(y|c)$, indicating that a concept is strongly associated with only a few classes.
\medbreak
\noindent \textbf{Coverage Score.} The second term of equation \ref{eq: submodular function} is a minimax facility location function that tries to minimize the maximum distance between each element in the subset and the candidate set. For distance, we use the cosine between the features of the two concepts extracted by the text encoder: $\phi(c_1, c_2) = \text{cos} \left(\mathcal{T}(c_1), \mathcal{T}(c_2)\right)$.
A high coverage score yields a diverse bottleneck that covers different possible appearances for a target class. 

\subsection{Optimize Class-concept Association} \label{sec: linear function}
In this section, we explain how we compute $g$ (the concept predictor) and learn $f$ (the label predictor) of the bottleneck.
\medbreak
\noindent \textbf{Predict the Concept Scores.} The concept predictor $g$ is not learned in our method because the alignment model we use can measure the correlation between image and text through dot product. We treat the dot product of input image feature $\boldsymbol{x}$ and the concept space $\boldsymbol{E}_C$ defined in Section \ref{sec: concept selection} as $g$: $g\left(\boldsymbol{x}, \boldsymbol{E}_C\right) = \boldsymbol{x} \cdot \boldsymbol{E}_C^{\top}$, where $g\left(\boldsymbol{x},\boldsymbol{E}_C\right) \in \mathbb{R}^{N_C}$, and each element is the score of image $\boldsymbol{x}$ on a concept.
\medbreak
\noindent \textbf{Concept Weight Matrix.} 
We learn a linear function for the label predictor $f$ that maps from concept scores to the final prediction.
Intuitively, these weights encode the affinity of the concept to the class, allowing the model to represent that classes depend differently on the same concept.
To normalize the class-concept association distributed over the weight matrix, we regularize the matrix with the softmax activation function.
Concretely, we learn a concept weight matrix $\boldsymbol{W} \in \mathbb{R}^{N \times N_C}$, that is used for prediction: 
$
    \hat{y} = \text{argmax} \left(g\left(\boldsymbol{x}, \boldsymbol{E}_C\right) \cdot \sigma\left(\boldsymbol{W}\right)^{\top}\right),
$
in which $\sigma(\cdot)$ is the softmax activation which is applied along the concepts axis:
$\boldsymbol{W}_{y,c} = e^{\boldsymbol{W}_{y,c}}/\sum_{y' \in \mathcal{Y}} e^{\boldsymbol{W}_{y',c}}$.
\medbreak
\noindent \textbf{Initializing the Weight Matrix with Language Priors.} Previous work trains the concept weight matrix freely from scratch, which is not feasible in low-resource scenarios where we don't have enough data to learn the weight effectively. To extend the application of CBM to few-shot image classification, we consider biasing the weights toward the initial association from the language model used to propose concepts.
If a concept $c$ was present in $C_y$, we initialize the elements of $\boldsymbol{W}$ corresponding to the weight between class $y$ and concept $c$ to a higher value before optimization: $\boldsymbol{W}_{y,c} = 1$, $\text{if $c \in C_y$}$, otherwise 0.
\begin{figure*}[!t]
\centering
  \begin{subfigure}{0.24\textwidth}
    \centering
    \includegraphics[width=.99\linewidth]{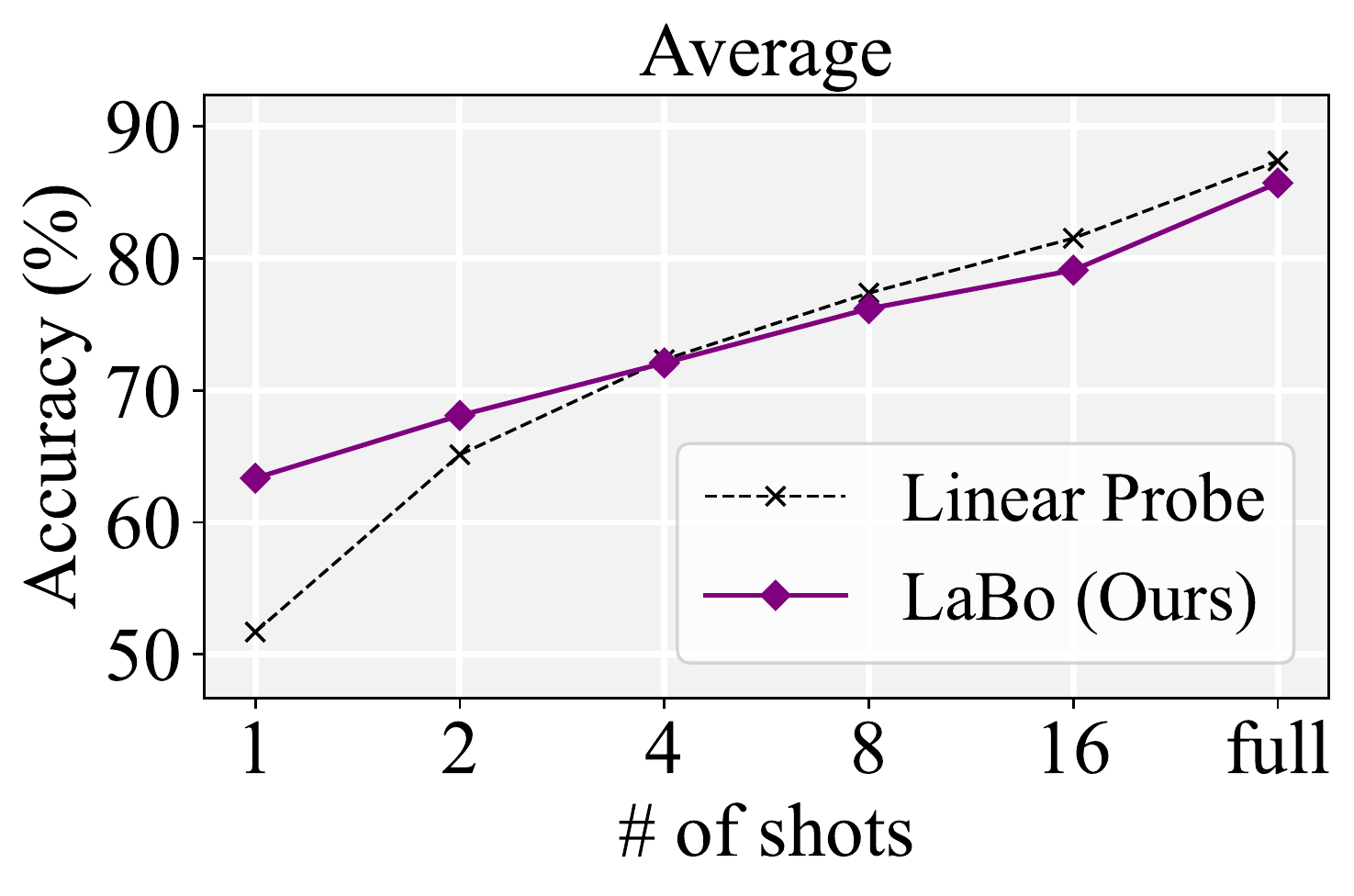}
  \end{subfigure}
  \begin{subfigure}{0.24\textwidth}
    \centering
    \includegraphics[width=.99\linewidth]{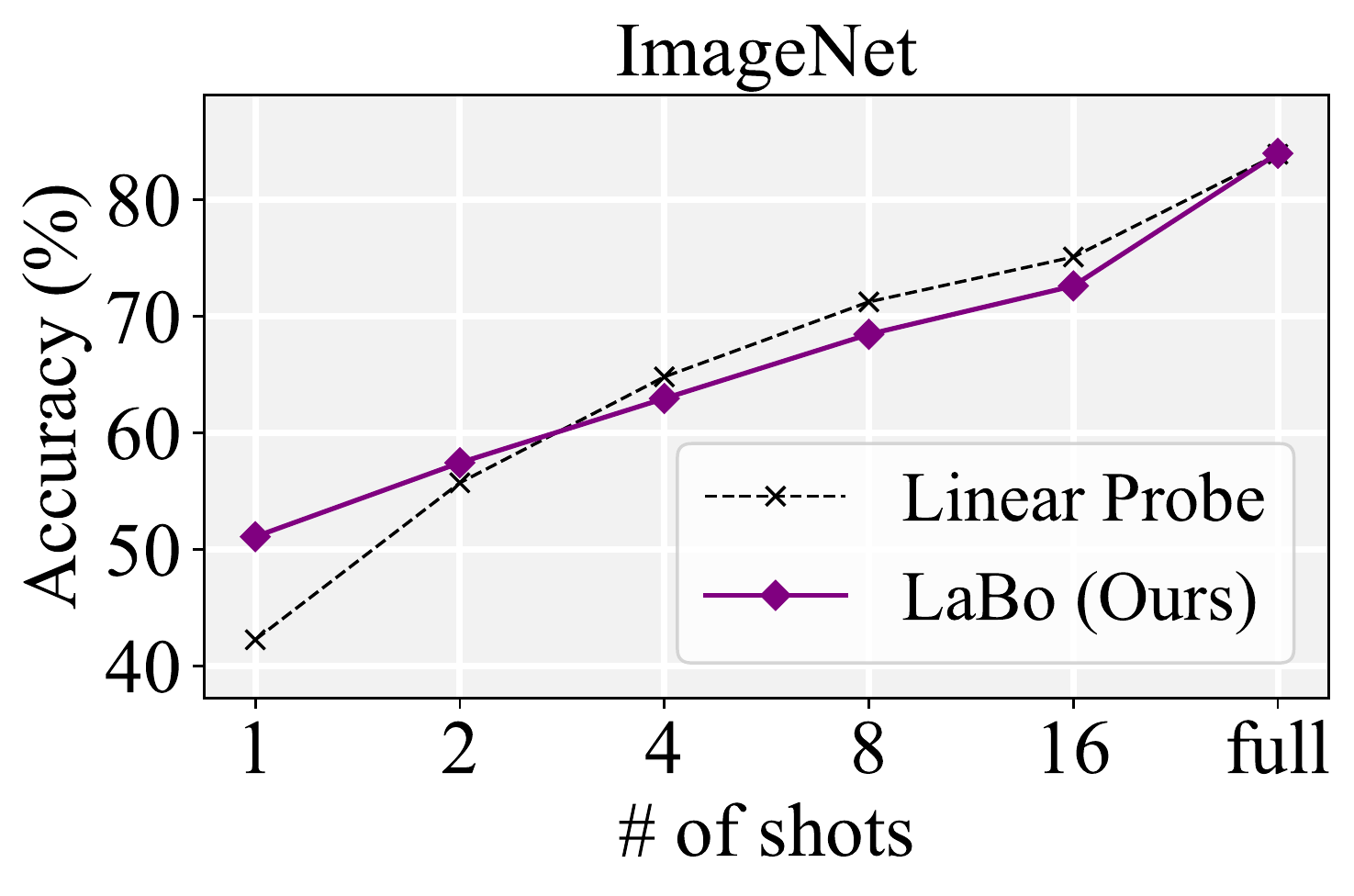}
  \end{subfigure}%
  \begin{subfigure}{0.24\textwidth}
    \centering
    \includegraphics[width=.99\linewidth]{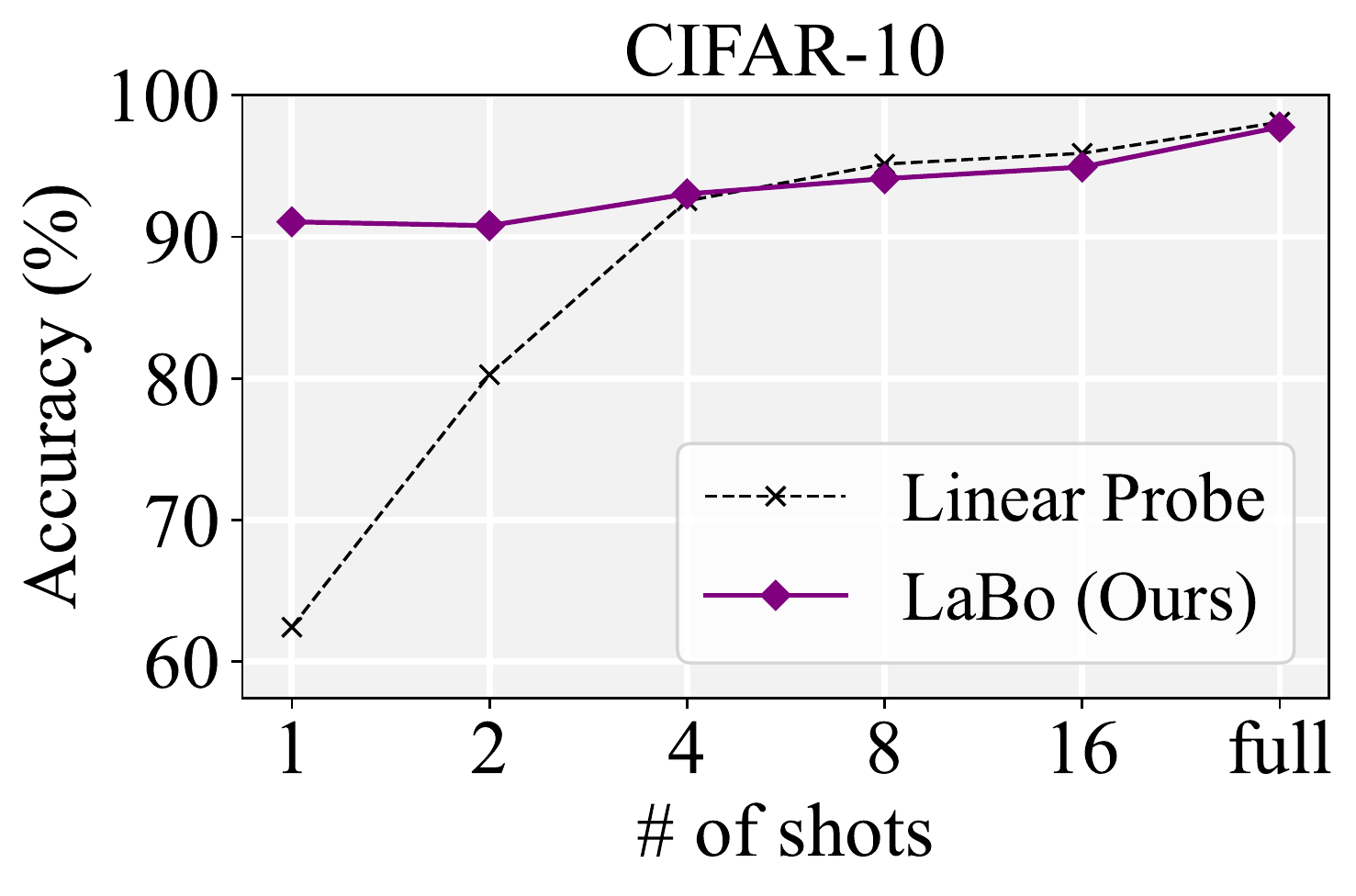}
  \end{subfigure}
  \begin{subfigure}{0.24\textwidth}
    \centering
    \includegraphics[width=.99\linewidth]{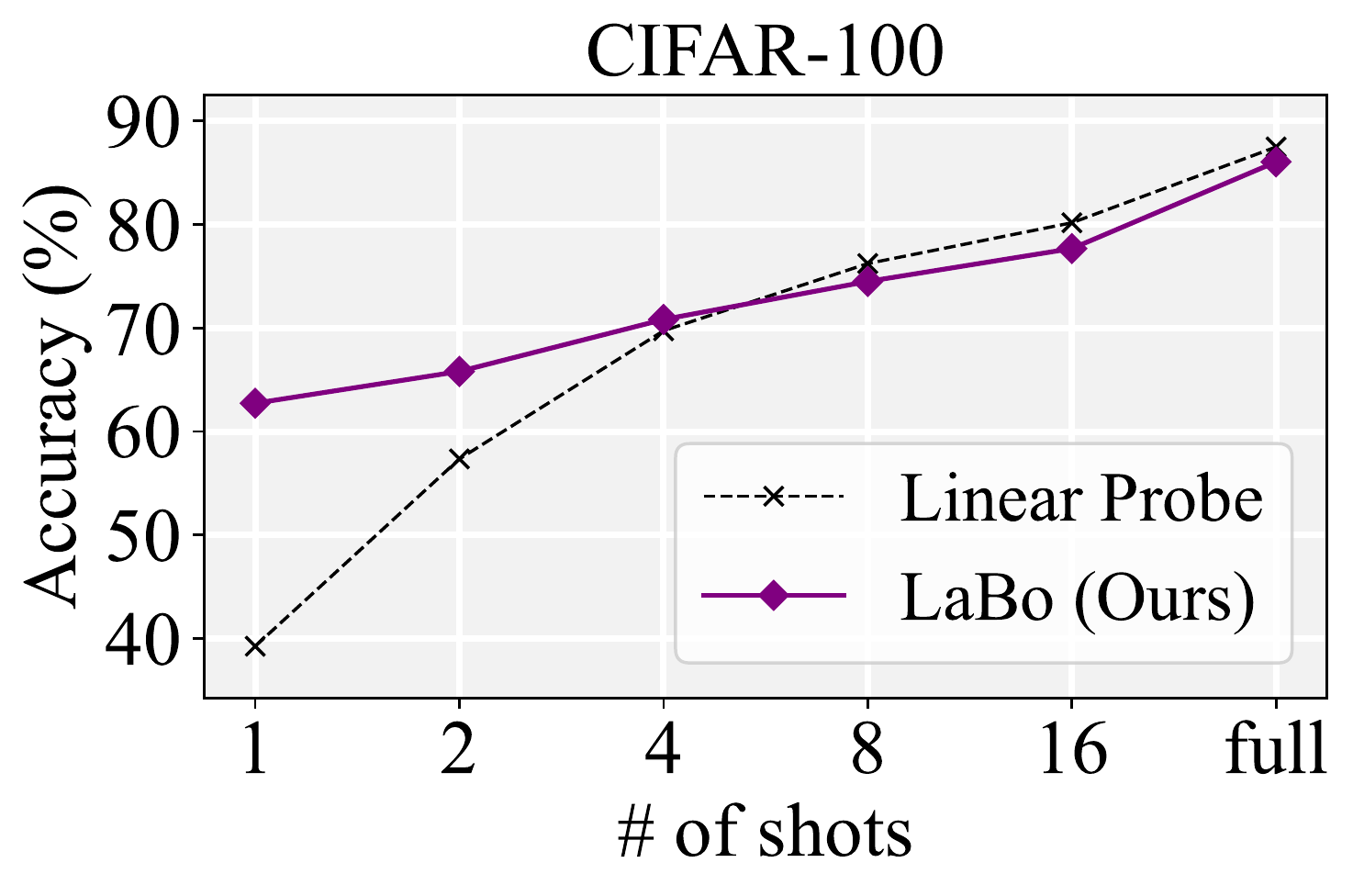}
  \end{subfigure}
  \begin{subfigure}{0.24\textwidth}
    \centering
    \includegraphics[width=.99\linewidth]{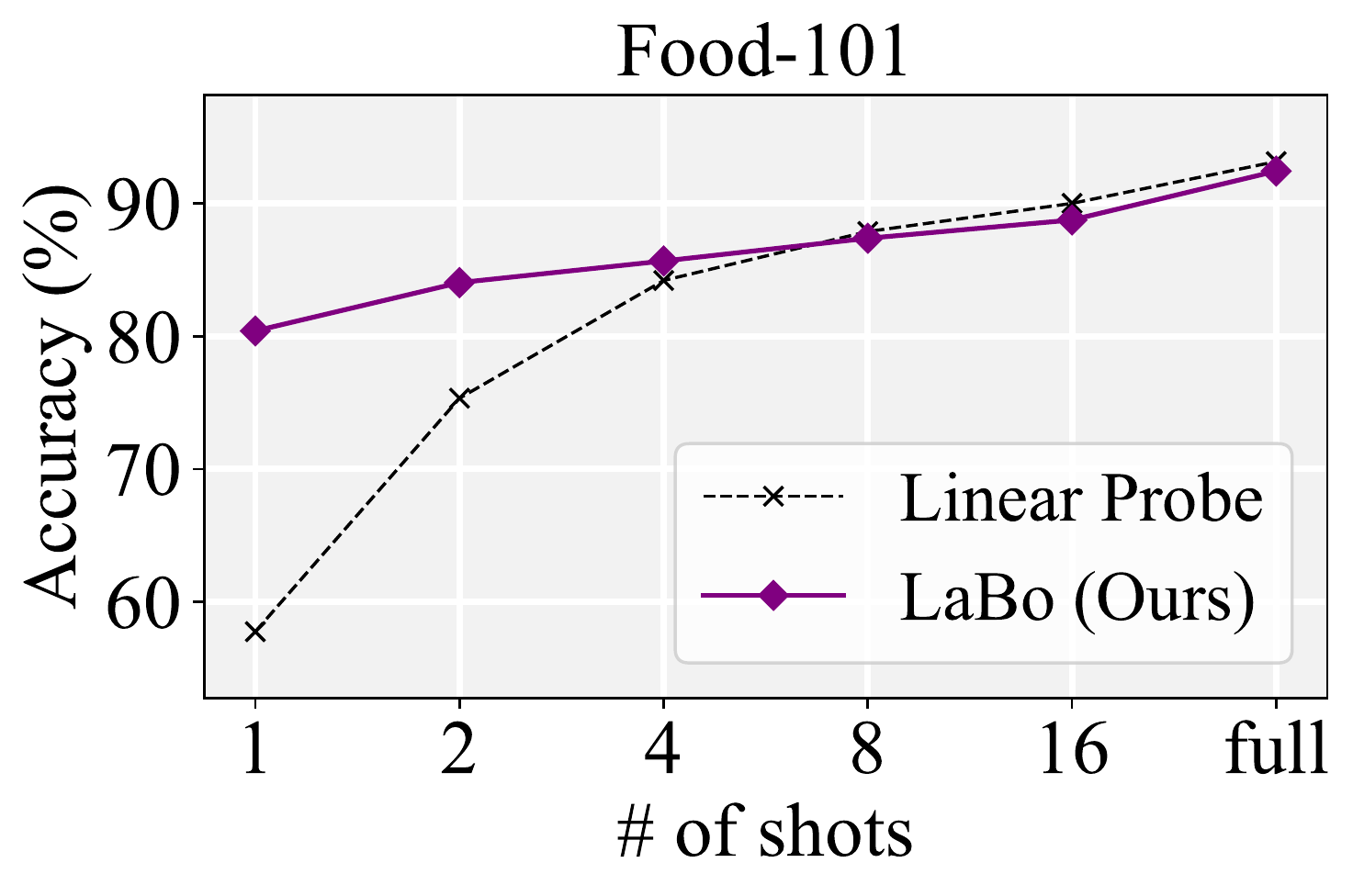}
  \end{subfigure}
  \begin{subfigure}{0.24\textwidth}
    \centering
    \includegraphics[width=.99\linewidth]{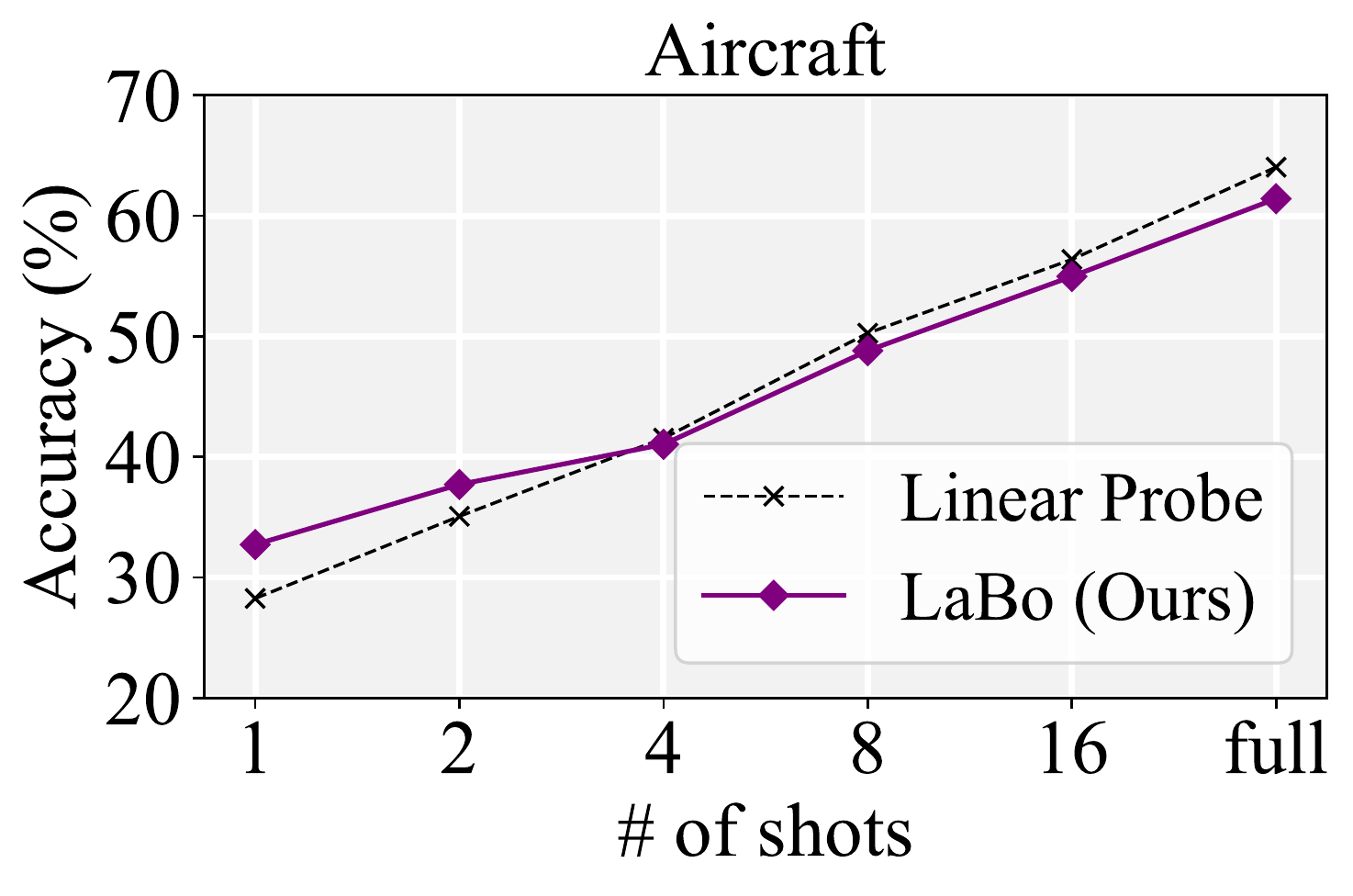}
  \end{subfigure}
  \begin{subfigure}{0.24\textwidth}
    \centering
    \includegraphics[width=.99\linewidth]{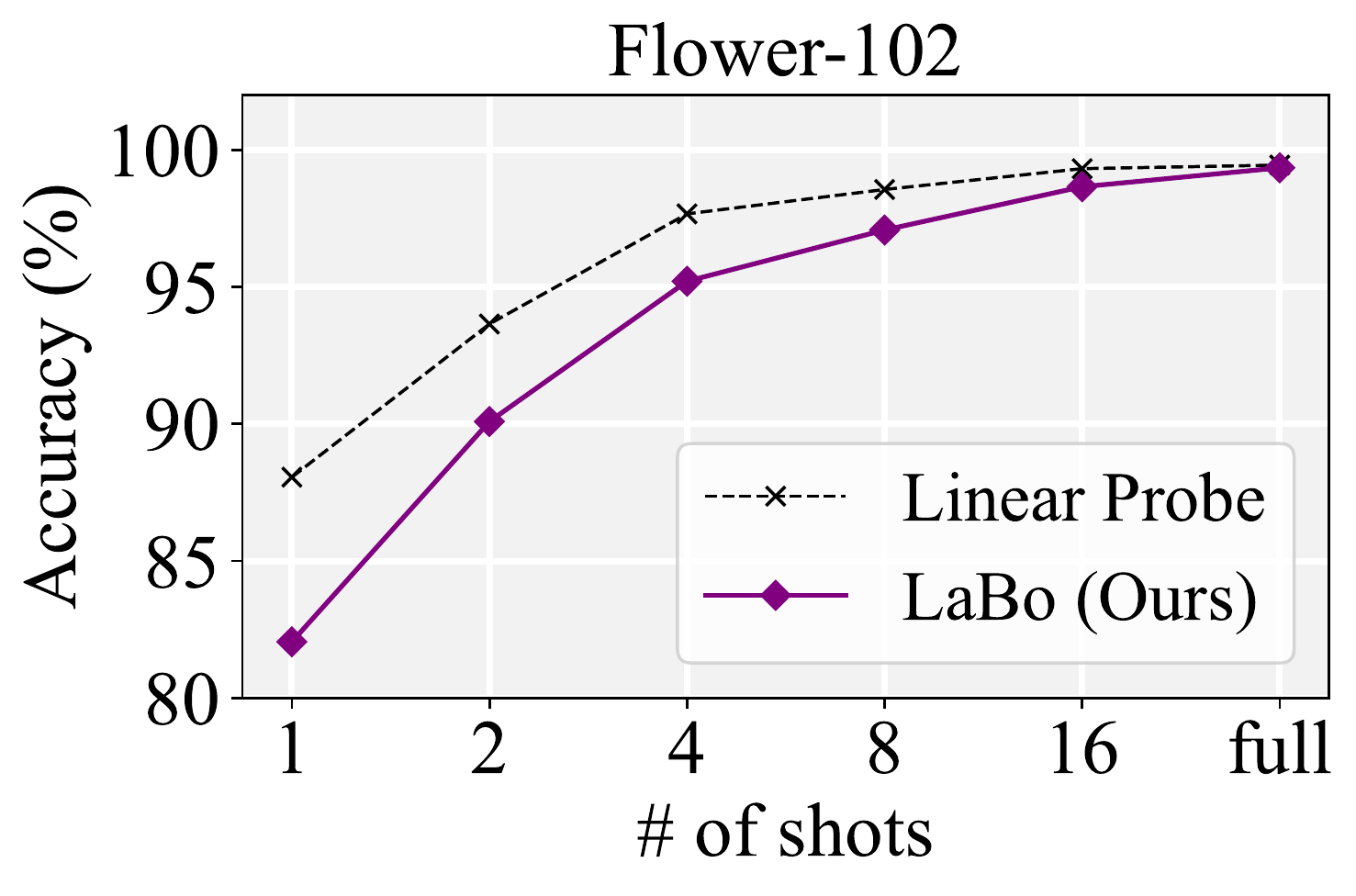}
  \end{subfigure}
  \begin{subfigure}{0.24\textwidth}
    \centering
    \includegraphics[width=.99\linewidth]{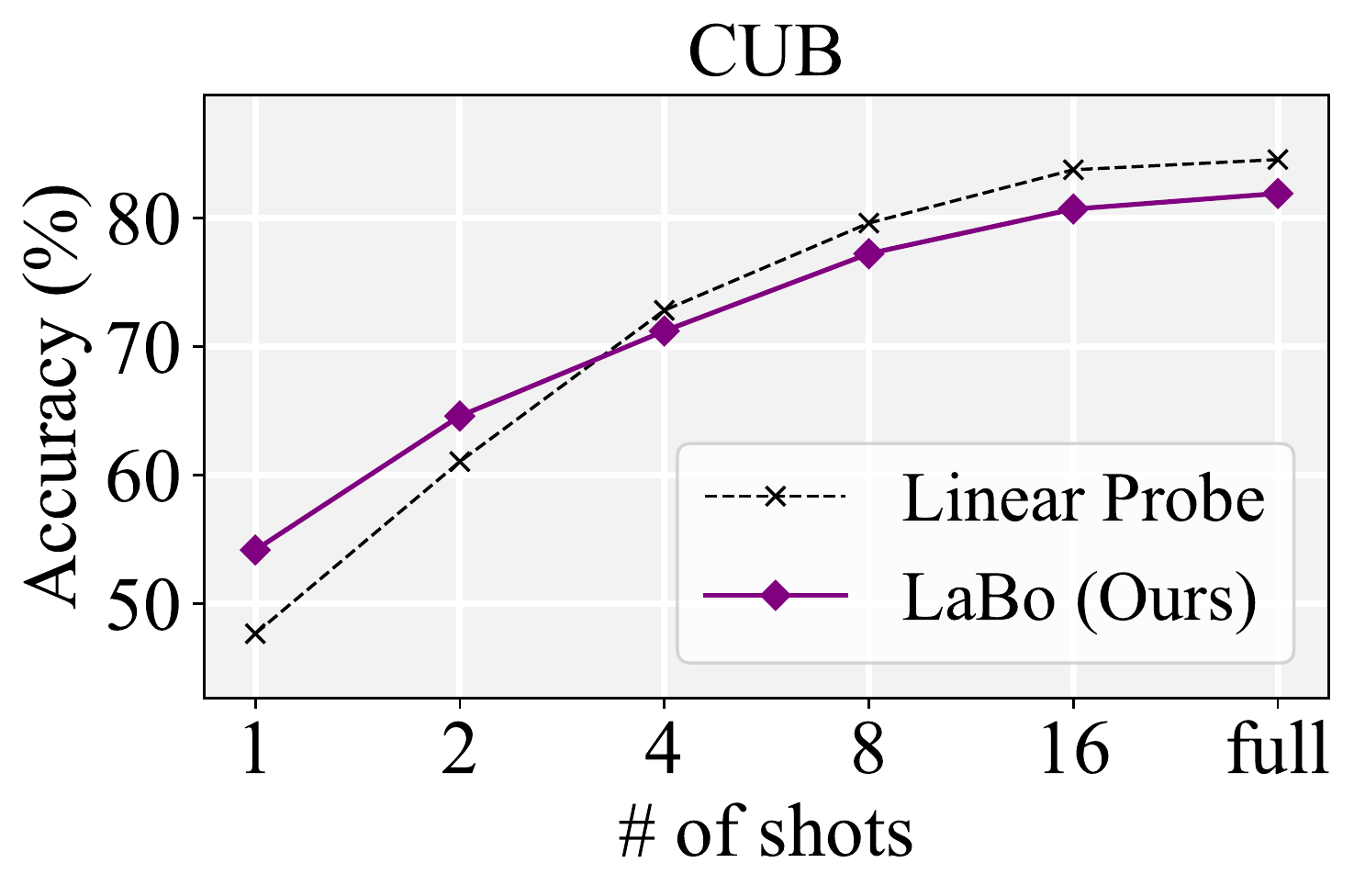}
  \end{subfigure}
  \begin{subfigure}{0.24\textwidth}
    \centering
    \includegraphics[width=.99\linewidth]{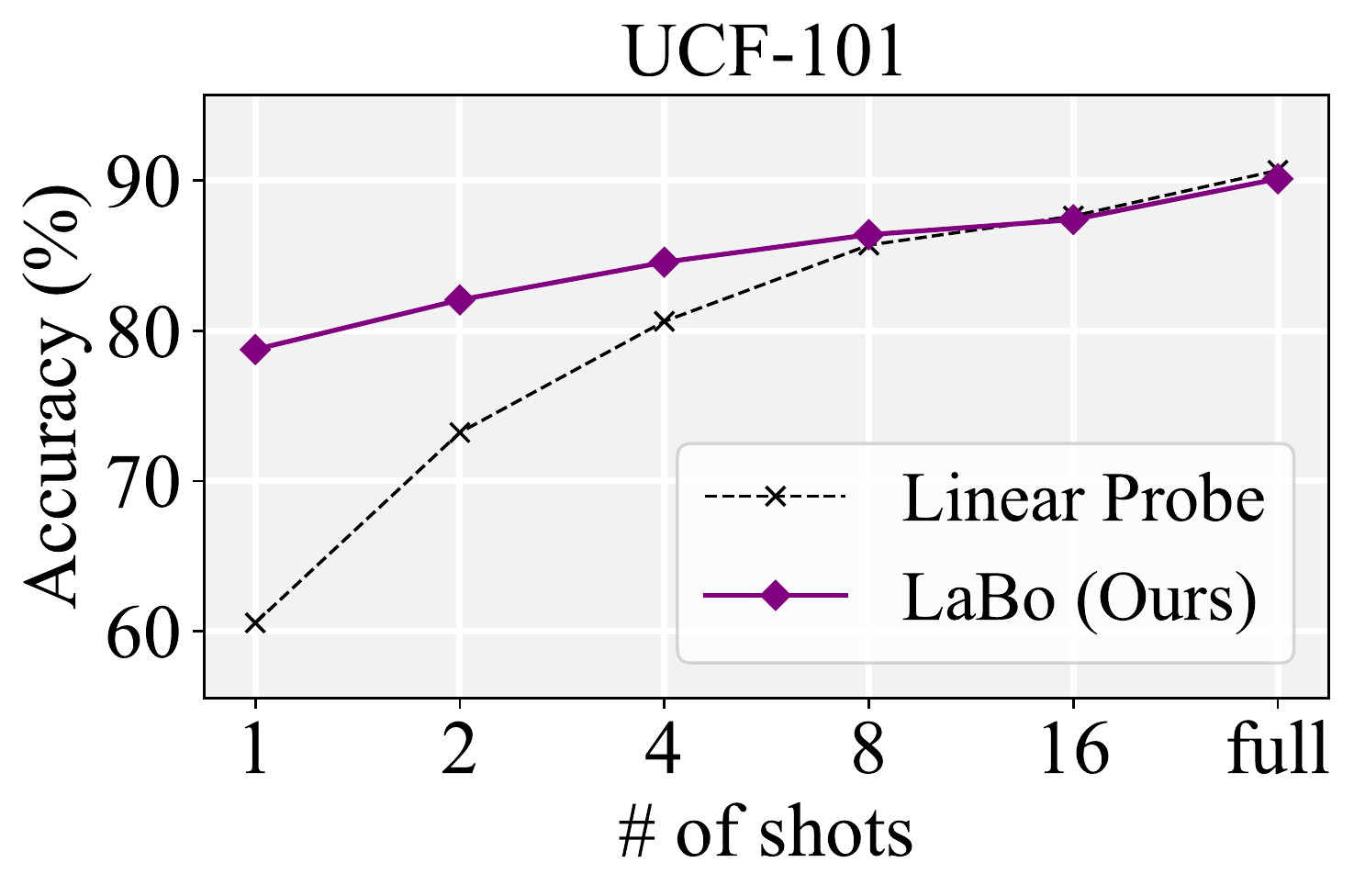}
  \end{subfigure}
  \begin{subfigure}{0.24\textwidth}
    \centering
    \includegraphics[width=.99\linewidth]{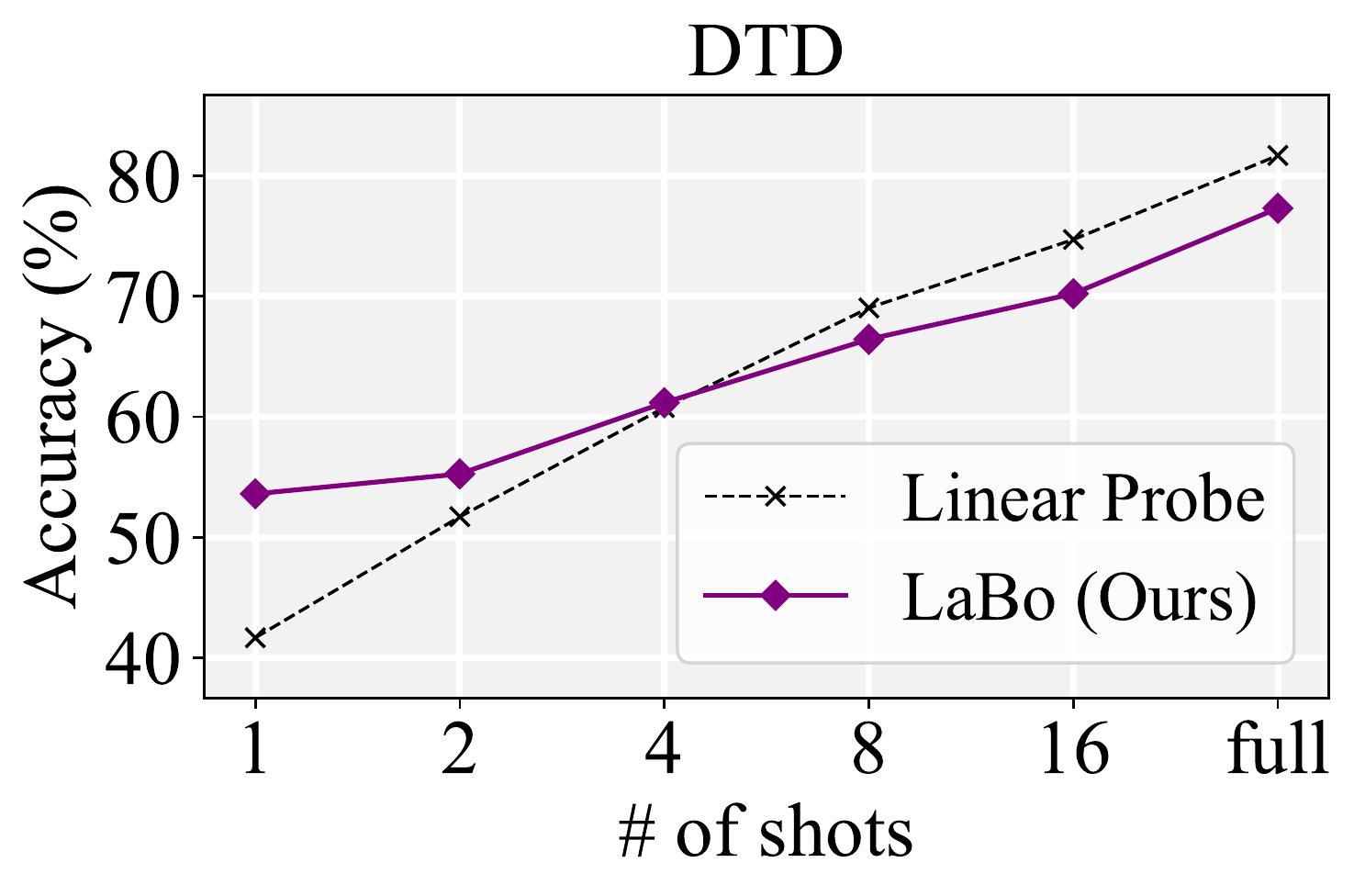}
  \end{subfigure}
  \begin{subfigure}{0.24\textwidth}
    \centering
    \includegraphics[width=.99\linewidth]{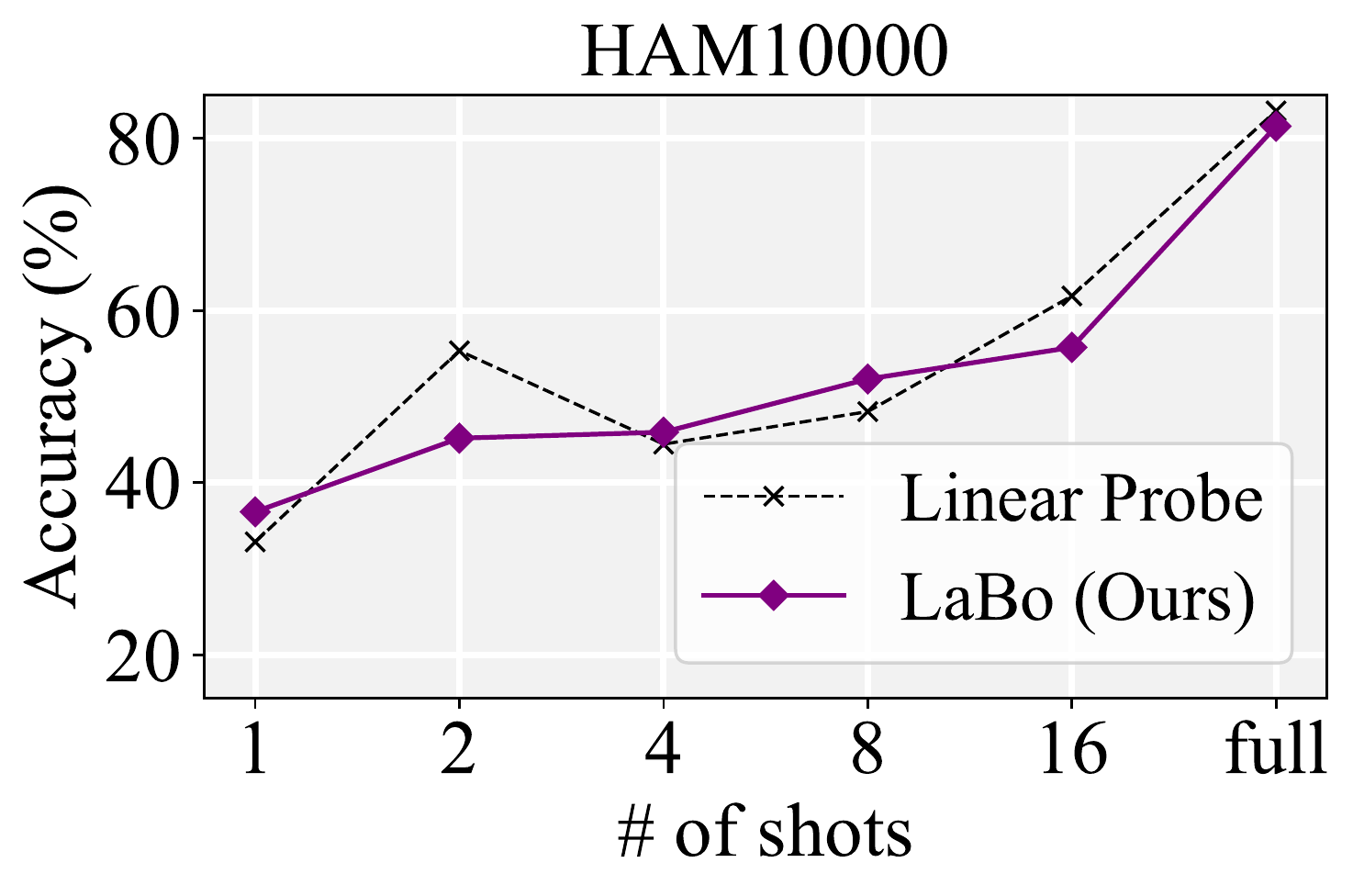}
  \end{subfigure}
  \begin{subfigure}{0.24\textwidth}
    \centering
    \includegraphics[width=.99\linewidth]{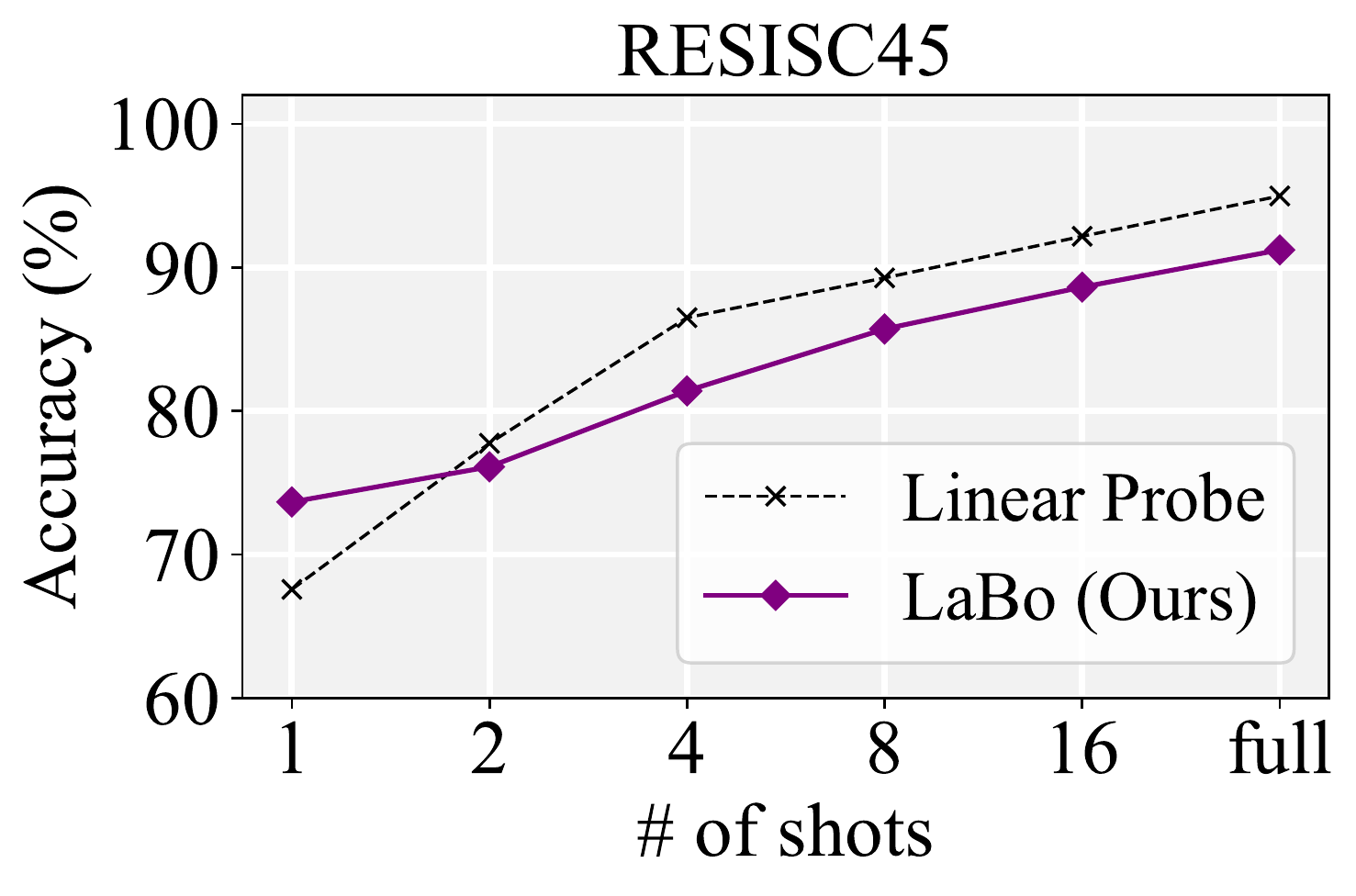}
  \end{subfigure}

  \caption{Test accuracy (\%) comparison between LaBo and Linear Probe on 11 datasets. The x-axis represents the number of labeled images.}
  \label{fig: main results}
  \vspace{-0.3cm}
\end{figure*}

\subsection{Prepare the Candidates} \label{sec: generation}
To collect the candidates $S$ to feed into our model, we prompt GPT-3 to generate relevant sentences by incorporating the class name in 5 templates shown in supplementary materials.\footnote{We use the same set of prompts for all datasets except UCF-101 since it is very different to describe an action.} For example, as shown in the top-left of Figure \ref{fig: diagram}, we prompt GPT-3 by asking ``\textit{describe what the \textbf{axolotl} looks like}'', and the GPT-3 returns a sentence about the target class. We obtain 500 sentences for each class and automatically split these sentences into shorter concepts using a T5 model \cite{2020t5} fine-tuned on a small set of annotated sentence-concept pairs. We use string match to identify and remove class name tokens in each concept. (see supplementary)

\section{Experimental Setup}
We evaluate our method on a diverse set of 11 datasets (Section \ref{sec:datasets}) and compare it to its end-to-end counterpart and other interpretable CBM methods (Section \ref{sec:baselines}).  

\subsection{Dataset}
\label{sec:datasets}
We select a comprehensive benchmark of 11 image classification datasets spanning a diverse set of domains, including (1) Common objects: ImageNet \cite{deng2009imagenet}, CIFAR-10 and CIFAR-100 \cite{krizhevsky2009learning}; (2) Fine-grained objects: Food-101 \cite{bossard14}, FGVC-Aircraft \cite{maji2013fine}, Flower-102 \cite{Nilsback08}, CUB-200-2011 \cite{wah2011caltech}; (3) Actions: UCF-101 \cite{soomro2012ucf101}; (4) Textures: DTD \cite{cimpoi2014describing}; (5) Skin tumors: HAM10000 \cite{tschandl2018ham10000} and (6) Satellite images: RESISC45 \cite{cheng2017remote}. 
We use train/dev/test splits
for all the datasets. Detailed statistics are presented in the supplementary material. 
We follow the few-shot evaluation protocol proposed by CLIP \cite{radford2021learning} with 1, 2, 4, 8, and 16 images randomly sampled from the training set for each class. We also evaluate in the fully-supervised setting where we train on all available images. For all experiments, we report the test accuracy.

\subsection{Baselines}
\label{sec:baselines}
We compare our model, LaBo, with black-box linear probing and two interpretable methods.
\medbreak
\noindent \textbf{Linear Probe} Following previous evaluations on CBM~\cite{koh2020concept, yuksekgonul2022post}, linear probing serves as our primary baseline for comparison. 
We follow the implementation of CLIP \cite{radford2021learning} by training the scikit-learn's L-BFGS logistic regression with a hyperparameter sweep on the L2 regularization weight.
\medbreak
\noindent \textbf{PCBM} Post-hoc Concept Bottleneck Model \cite{yuksekgonul2022post} designs a residual modeling step that directly maps the original image embedding into the label space. PCBM treats the attributes of each class in ConceptNet \cite{speer2017conceptnet} as concepts. 

\medbreak
\noindent \textbf{CompDL} Compositional Derivation Learning \cite{yun2022vision} learns a linear layer over CLIP similarity scores between human-designed class descriptions and images to predict targets. 

\subsection{Implementation Details}
\label{sec:implementation_details}
We prompt \href{https://beta.openai.com/docs/models/gpt-3}{GPT-3-text-davinci-002} to generate concepts.
The CLIP model is adapted from \href{https://github.com/openai/CLIP}{OpenAI's public repo}
with ViT-L/14 as the default vision backbone. We only use CLIP-RN50 as the backbone when comparing with PCBM, and ViT-B/32 with CompDL for a fair comparison. We implement the submodular function using the \href{https://github.com/jmschrei/apricot}{apricot package}
and set the default number of concepts selected for each class to 50. To train the linear function, we use the \href{https://www.pytorchlightning.ai/}{Pytorch-lightning library}
with Adam \cite{kingma2014adam} optimizer. 
We tune the batch size, learning rate, and submodular weights on the development set. 
Model checkpoints with the highest validation accuracy are evaluated on the test set. We list the hyperparameters for all datasets and shots in the supplementary material.

\section{Evaluation}
\subsection{Main Results}
We compare LaBo's performance with a linear probe and other interpretable baselines to evaluate if we can maintain black box accuracy without sacrificing interpretability.
\medbreak
\noindent \textbf{Comparison with End-to-End Model.} One of our goals is to close the performance gap between interpretable and black box models.
Table \ref{tab: main number} reports the mean test accuracy of LaBo and the linear probe on 11 datasets. 
LaBo significantly outperforms the end-to-end model when little data is available and continues to be competitive as the number of data increases.
On average, LaBo surpasses the linear probe by 1.5\%. 
Figure \ref{fig: main results} provides analytic performance comparisons between LaBo and Linear Probe on each dataset.  

\begin{table}[!t]
\centering
\resizebox{8.3cm}{!}{%
\begin{tabular}{cccccccc} \Xhline{3\arrayrulewidth}  
\textbf{Method}      & \textbf{1}      & \textbf{2}      & \textbf{4}      & \textbf{8}      & \textbf{16}     & \textbf{Full} & \textbf{Avg}  \\ \hline
Linear Probe &   51.69 & 65.13 & \textbf{72.33} & \textbf{77.38} & \textbf{81.53} & \textbf{87.38}   &  72.57  \\
LaBo (Ours) &   \textbf{63.35} & \textbf{68.10} & 72.08 & 76.19 & 79.11 & 85.72   &  \textbf{74.09}   \\ \Xhline{3\arrayrulewidth}  
\end{tabular}
}
\caption{Mean accuracy across all datasets, at different shots .}
\label{tab: main number}
\vspace{-0.1cm}
\end{table}
\begin{table}[!t]
\centering
\resizebox{7.5cm}{!}{%
\begin{tabular}{cccc}
\Xhline{3\arrayrulewidth}  
 \textbf{Method}           & \textbf{w/ end-to-end} & \textbf{CIFAR-10} & \textbf{CIFAR-100} \\ \hline

PCBM \cite{yuksekgonul2022post}       &    \xmark   &   84.5       &    56.0       \\
LaBo (Ours)        &    \xmark         &  \textbf{87.9}        &   \textbf{69.1}        \\ \hline
PCBM-h \cite{yuksekgonul2022post}     &    \cmark         &  87.6        &  69.9   \\  
Linear Probe &   \cmark         &  88.8        &   70.1        \\\Xhline{3\arrayrulewidth}   
\end{tabular}
}
\caption{Test accuracy comparison between LaBo and Post-hoc Concept Bottleneck Model (PCBM) on CIFAR-10 and CIFAR-100. ``w/ end-to-end'' denotes whether the model employs an end-to-end residual predictor from image features to targets.}
\label{tab: PCBM}
\vspace{-0.1cm}
\end{table}
\begin{table}[!t]
\centering
\resizebox{7.5cm}{!}{%
\begin{tabular}{ccccc} 
\Xhline{3\arrayrulewidth}  
\textbf{Method}    & \textbf{w/ manual concepts} & \textbf{1} & \textbf{5} & \textbf{Full} \\ \hline
CompDL \cite{yun2022vision} & \cmark   & 13.6   & 33.2  &  52.6     \\
LaBo (Ours)      & \xmark      & \textbf{35.1}  & \textbf{55.7}  &  \textbf{71.8} \\\hline
Linear Probe & -      & 28.4  & 55.4  & 75.5     \\\Xhline{3\arrayrulewidth}  
\end{tabular}
}
 \caption{LaBo and CompDL evaluated on CUB for 1/5/full shots.}
 \label{tab: compdl}
 \vspace{-0.3cm}
\end{table}
In general, LaBo's performance depends on the quality of knowledge extracted from GPT-3.
For common categories, GPT-3 contains high-quality knowledge allowing  substantial improvement over linear probes.
For some fine-grained datasets, such as Flower-102, GPT-3's knowledge is largely non-visual, as seen in Figure~\ref{fig: qual examples}.
In such cases, specialized language models could be used to improve LaBo. 
\medbreak
\noindent \textbf{Comparison with other Interpretable Methods.} 
Table \ref{tab: PCBM} compares LaBo's performance with PCBM and Linear Probe.
LaBo outperforms PCBM by 3.4\% on CIFAR-10 and 13.1\% on CIFAR-100.
LaBo maintains comparable performance to PCBM with a residual predictor (PCBM-h), without circumventing the bottleneck.
In Table \ref{tab: compdl}, LaBo is more accurate than CompDL\cite{yun2022vision} without manually constructed concepts.


\subsection{Ablation Study}\label{sec:ablation}
\begin{table}[!t]
\centering
\resizebox{8cm}{!}{%
\begin{tabular}{c|cccccc}
\Xhline{4\arrayrulewidth}  
\multirow{2}{*}{\begin{tabular}[c]{@{}c@{}}\textbf{n. of concepts}\\\textbf{per class} ($k$)\end{tabular}} & \multicolumn{6}{c}{\textbf{n. of shots}}                                  \\ \cline{2-7} 
& \textbf{1}     & \textbf{2}     & \textbf{4}     & \textbf{8}     & \textbf{16}    & \multicolumn{1}{l}{\textbf{Full}} \\ \hline
1     & 41.89 & 52.45 & 61.76 & 65.99 & 69.61 & 78.95    \\
5     & 52.54 & 61.13 & 67.22 & 72.90 & 75.62 & 83.83    \\
10    & 58.00 & 64.59 & 69.90 & 74.50 & 77.43 & 84.66    \\
25    & 61.72 & 66.33 & 71.39 & 75.28 & 79.04 & 85.26    \\
50    & \textbf{63.03} & \textbf{67.79} & \textbf{71.88} & \textbf{76.08} & \textbf{79.10} & \textbf{85.71} \\\Xhline{4\arrayrulewidth}  
\end{tabular}
}
\caption{Ablation results on bottleneck sizes. 
We vary the sizes of the bottlenecks
and report the average performance on 11 datasets.
}
\label{tab: bottleneck size}
\vspace{-0.1cm}
\end{table}

\begin{table}[!t]
\centering
\resizebox{8cm}{!}{%
\begin{tabular}{c|cccccc}
\Xhline{4\arrayrulewidth}  
\multirow{2}{*}{\begin{tabular}[c]{@{}c@{}}\textbf{Selection}\\\textbf{Method} \end{tabular}} & \multicolumn{6}{c}{\textbf{n. of shots}}                                  \\ \cline{2-7} 
& \textbf{1}     & \textbf{2}     & \textbf{4}     & \textbf{8}     & \textbf{16}    & \multicolumn{1}{l}{\textbf{Full}} \\ \hline
\textsc{Random}          & 59.24 & 64.71 & 70.42 & 74.07 & 78.29 & 85.06    \\
\textsc{Similarity}      & 54.59 & 61.42 & 67.17 & 72.66 & 77.32 & 84.88    \\
\textsc{Coverage}        & 59.73 & 65.93 & 70.82 & 74.71 & 78.90 & 85.60    \\
\textsc{Discrim}         & 60.99 & 66.49 & 70.93 & 74.81 & 77.90 & 85.31    \\
\textsc{Submodular}      & \textbf{63.03} & \textbf{67.79} & \textbf{71.88} & \textbf{76.08} & \textbf{79.10} & \textbf{85.71}   \\\Xhline{4\arrayrulewidth}  
\end{tabular}
}
\caption{Ablation results on concept selection methods. We report mean test accuracy on 11 datasets.}
\label{tab: selection methods}
\vspace{-0.3cm}
\end{table}
We evaluate the importance of each of our model's components on final performance. Specifically, we compare results with different concept selection methods, language and random weight initialization, and bottleneck sizes. 

\begin{figure}[!t]
    \includegraphics[width=8.3cm]{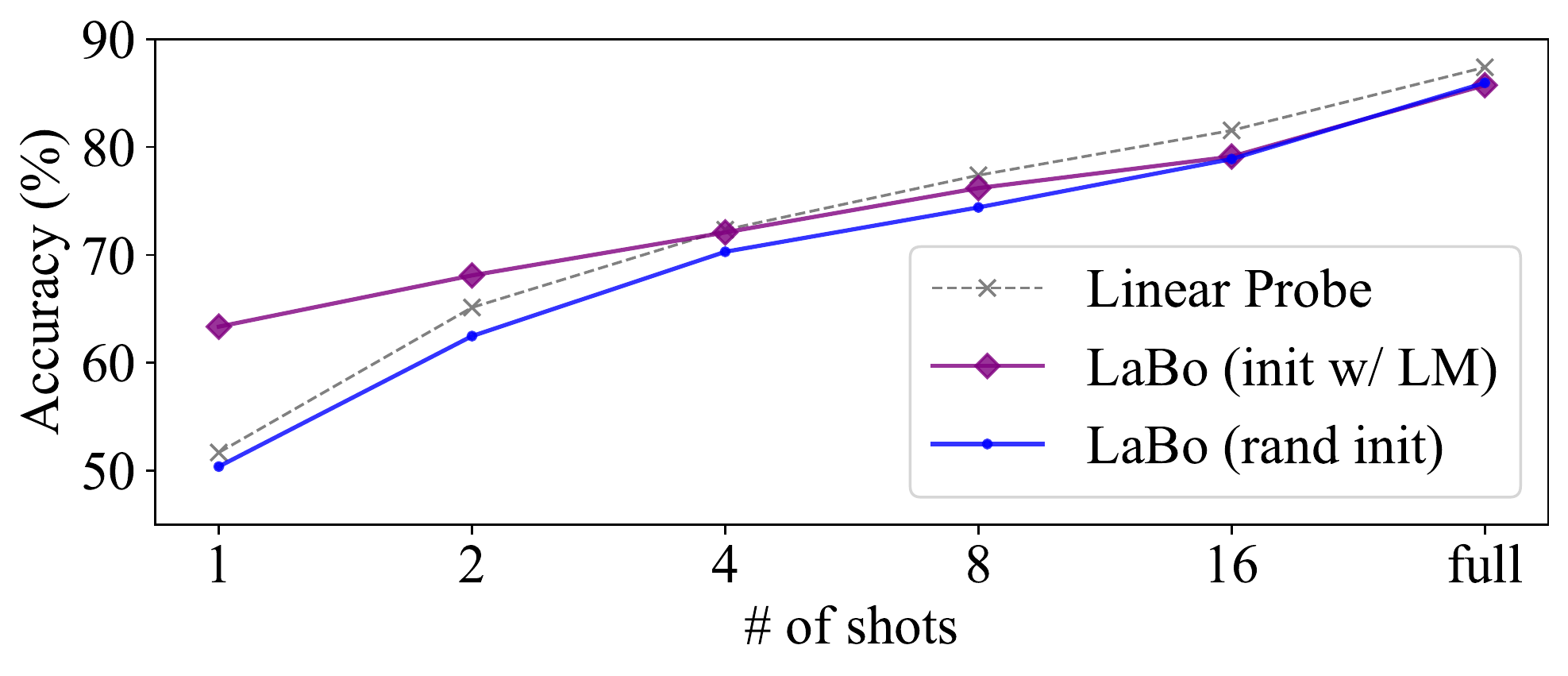}
    \caption{Language Prior vs. Random Weight initialization average over all datasets.
    }
    \label{fig: lm init}
    \vspace{-0.3cm}
\end{figure}

\medbreak
\noindent \textbf{Concept Selection Methods.} We compare our submodular function with four concept selection methods: (1) \textsc{Random}: we randomly sample a subset of concepts from the candidates for each class; (2) \textsc{Similarity}: we select the top concepts ranked by their similarity scores with the class calculated by equation \ref{eq; similarity score}; (3) \textsc{Coverage}: we only consider  the coverage score for concept selection; (4) \textsc{Discrim}: we only consider the discriminability score for concept selection. As shown in Table \ref{tab: selection methods}, our submodular function, which jointly optimizes coverage and discriminability, achieves the best performance across different numbers of shots. We notice that using coverage or discriminability alone 
still outperforms using similarity between the class and random selection. 
The selection method plays an important role in all data settings, but its impact decreases with more supervision.
\medbreak
\noindent \textbf{Initialization with Language Priors.} We deactivate the LM initialization and use random initialization instead. Figure \ref{fig: lm init} shows
that the LM prior is more important for low shot settings since there is less signal to guide concept importance. 

\medbreak
\noindent \textbf{Bottleneck Size.} 
In Table~\ref{tab: bottleneck size}, we compare performance for different bottleneck sizes ranging from 1 to 50 concepts selected by the submodular function.
Larger bottlenecks are usually better, but with more data, similar performance is achievable with smaller bottlenecks. 

\medbreak

\subsection{Human Evaluation}\label{sec:human_eval}

\begin{figure}[!t]
  \begin{subfigure}{0.47\textwidth}
    \includegraphics[width=.99\linewidth]{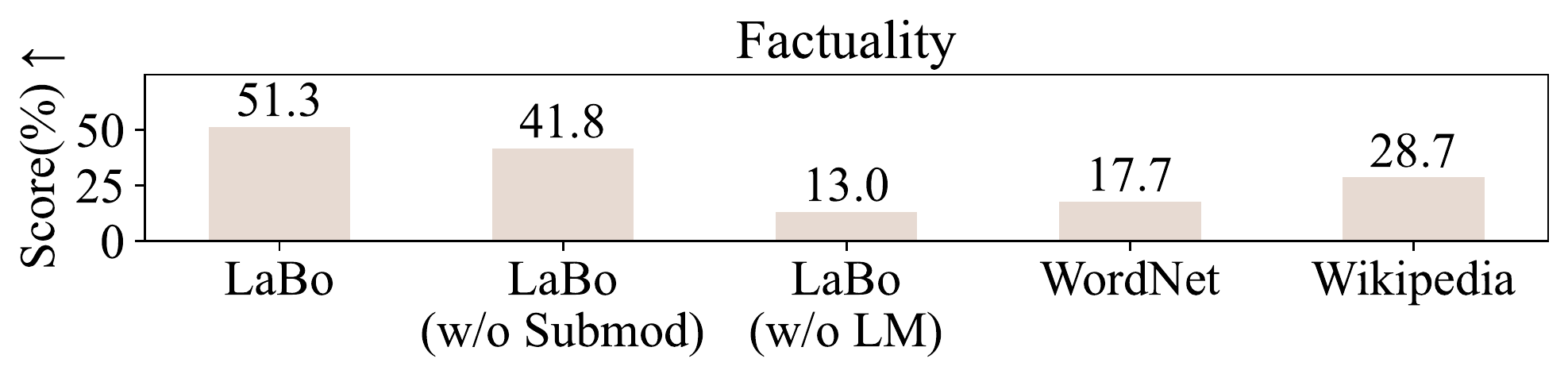}
  \end{subfigure}
  \begin{subfigure}{0.47\textwidth}
    \includegraphics[width=.99\linewidth]{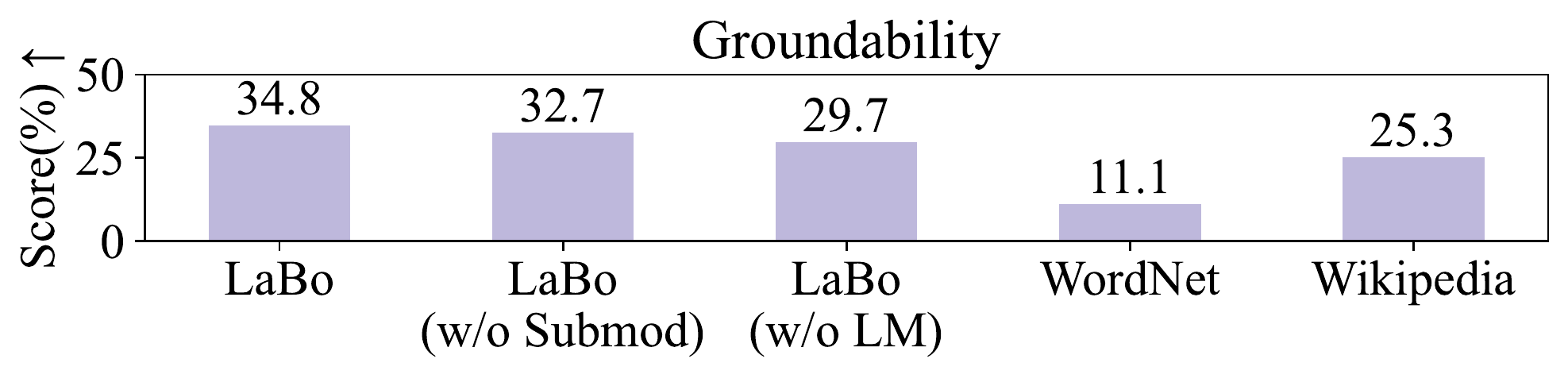}
  \end{subfigure}
  \caption{Human evaluation on \textit{Factuality} and \textit{Groundability} for different bottlenecks on ImageNet. ``w/o Submod" denotes without submodular function, i.e., random concept selection. ``w/o LM" denotes no language prior weight initialization.}
  \label{fig: imagenet human eval}
  \vspace{-0.3cm}
\end{figure}

It is important for interpretability that the vision-language alignment model correctly grounds concepts to images. For example, if a concept ``usually round" ranks both circles and stripes highly, the name of the attribute does not faithfully represent the computation. 
In addition, it is important that the automatically generated concept bottlenecks factually correspond to the class they describe. 
To this end, we introduce two metrics to evaluate the quality of our concept bottleneck items: (1) \textit{Factuality} measures how accurate the concepts are in describing their designated class by requiring annotators to judge whether they describe ground truth images, and (2) \textit{Groundability} measures how consistent the vision-language model grounding of the concepts to images with human interpretations by requiring annotators to judge their applicability on the top-10 images ranked by CLIP alignment scores. 
\medbreak
\noindent \textbf{Setup.} Both metrics are computed by asking annotators to select images that describe a highly ranked concept in our bottlenecks. Formally, the two metrics are represented by:
$$\textit{Factuality}(c) = \frac{\text{number of images selected}}{k \text{ ground truth images of the class}}$$
$$\textit{Groundability}(c) = \frac{\text{number of images selected}}{\text{top-\textit{k} aligned images of the concept}}$$
where we set $k=10$.\footnote{With the only exception of \textit{Factuality} for Flower-102 where we set $k=8$ because there are not enough images in the dev set.} In addition to the two main metrics, we ask the annotator to select whether the concept is non-visual, nonsensical, or contains unknown vocabulary. We randomly sample 20 classes for each dataset and evaluate the top 5 concepts (ranked by the weights of the linear function) for each class, 100 concepts per dataset. We release our human evaluation task on \href{https://www.mturk.com}{Amazon Mechanical Turk} and collect three annotations for each concept. More details on the task and the results can be found in the supplement. 
\medbreak

\begin{figure}[!t]
    \includegraphics[width=8.3cm]{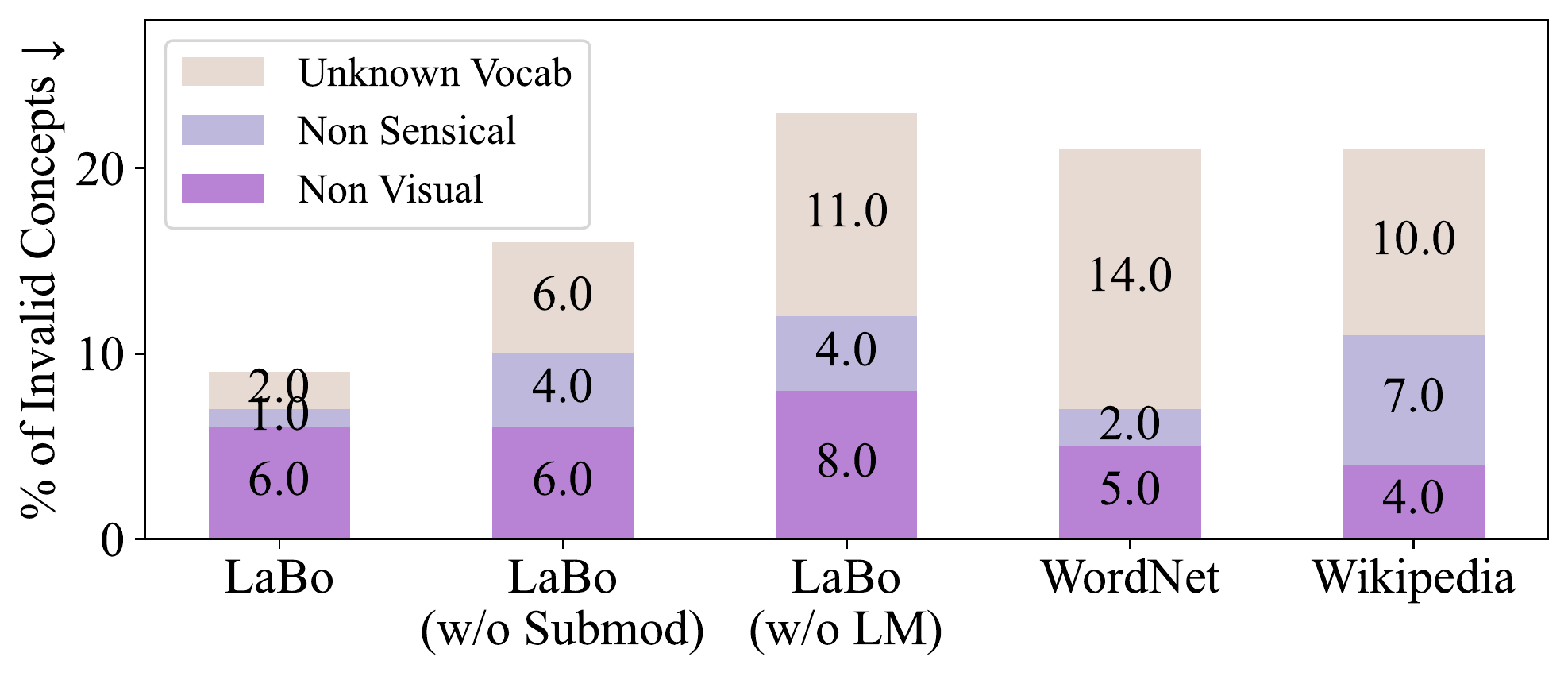}
    \caption{Percentage of invalid concepts identified by humans for different bottlenecks on ImageNet. \textbf{Lower} percentage is better.}
    \label{fig: invalid concepts}
    \vspace{-0.3cm}
\end{figure}

\begin{figure*}[!t]
\centering
    \includegraphics[width=17.3cm]{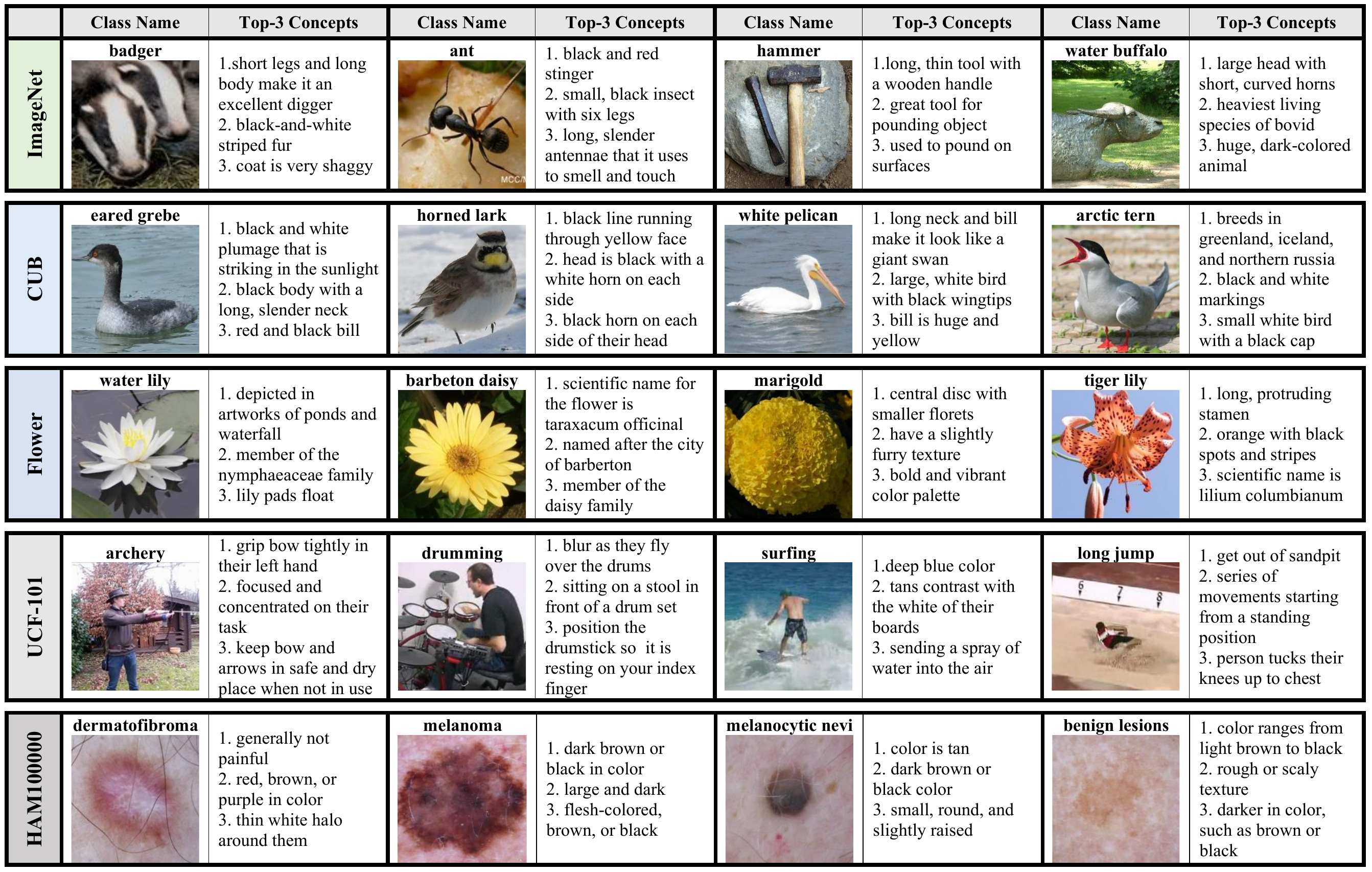}
    \caption{Several example bottlenecks generated by LaBo. The top-3 concepts, ranked by their weights in the linear function, for randomly selected classes, paired with a random image from the class, across 6 datasets. 
    }
    \label{fig: qual examples}
    \vspace{-0.3cm}
\end{figure*}

\begin{table}[!t]
\centering
\resizebox{8cm}{!}{%
\begin{tabular}{lccc}
\Xhline{3\arrayrulewidth}
\textbf{Metrics} & \textbf{LaBo} & \textbf{w/o Submod} & \textbf{w/o LM} \\ \hline
Factuality (\%) $\uparrow$   & \textbf{24.0} & 22.8	&14.1   \\
Groundability (\%) $\uparrow$ & 14.1 & \textbf{22.5}  & 20.2  \\
Non Visual (\%) $\downarrow$   & \textbf{4.8}  &	5.6 &  5.8       \\
Non Sensical (\%) $\downarrow$ & \textbf{8.0}	& 9.6 & 8.7     \\
Unknown Vocab (\%) $\downarrow$ & \textbf{10.2}  & 10.5  &	10.7 \\\Xhline{3\arrayrulewidth}    
\end{tabular}
}
\caption{Average human evaluation results of LaBo on 11 datasets. We also evaluate LaBo by removing the submodular function (w/o Submod) and language model priors (w/o LM).}
\label{tab: human_avg}
\vspace{-0.2cm}
\end{table}

\noindent \textbf{Baselines.} We evaluate the bottlenecks under full supervision and compare them with two main baselines: (1) LaBo (w/o Submod), which randomly selects the concepts instead of using the submodular function, and (2) LaBo (w/o LM), which initializes the concept weight matrix randomly without leveraging the priors of the language model. For ImageNet, we add two additional baselines using human-written text: (1) WordNet \cite{fellbaum2010wordnet} definitions and (2) Wikipedia sentences \cite{kil-chao-2021-revisiting}. We adopt the same preprocessing pipeline as LaBo to extract concepts from human-written resources and utilize the submodular function to select the bottlenecks.
\medbreak
\noindent \textbf{Results.} Figure \ref{fig: imagenet human eval} shows the evaluation on ImageNet, and we observe that LaBo has significantly higher \textit{Factuality} and \textit{Groundability} than  human-written text. We further observe that removing components from our system (submodular and LM Prior) hurt both human evaluation metrics, indicating their collective importance in our system. In addition, Figure \ref{fig: invalid concepts} shows that LaBo has significantly fewer invalid concepts than other baselines. Table \ref{tab: human_avg} summarizes the average human evaluation results over the 11 datasets\footnote{The low resolution of CIFAR images partially affects those metrics since annotators have greater difficulty in completing the task.
}. 
On average, we observe a trade-off between \textit{Factuality} and \textit{Groundability}. Increasing coverage and discriminability leads to more variable and specific concepts that CLIP finds more difficult to ground. 
This could be due to challenges in  
capturing composite concepts \cite{yun2022vision,mancini2021open}.
 For individual analysis of the datasets, refer to the supplementary material. 
 Finally, Figure \ref{fig: qual examples} shows several CBMs we constructed. 
 Across many types of tasks, the bottlenecks are largely coherent, factual, and groundable by CLIP.


\section{Conclusion and Limitation}
Overall, our approach demonstrates that the accuracy and interpretability of vision systems may be less at odds than previously believed.
Leveraging LLMs was crucial, as they encode important visual knowledge.
In the future, our approach can easily be enriched with new factors that capture different priors on bottleneck construction. 
The limits of knowledge in GPT-3 are not known, but likely there are domains where prompting generates few useful facts. 
Even in contexts where GPT-3 can generate useful information, our method depends on CLIP being able to recognize those aspects in images.
The alignment between GPT-3 and CLIP likely does not hold for all cases.
Future work could focus on dynamically prompting GPT-3 to make this coupling more robust.
Finally, our work depends on large models trained at scales that are not currently  reproducible. 
It is possible unrevealed aspects of training by OpenAI will require a reevaluation of our claims.

\section*{Acknowledgements}
This research is based upon work supported in part by the DARPA KAIROS Program (contract FA8750-19-2-1004), the DARPA LwLL Program (contract FA8750-19-2-0201), the IARPA BETTER Program (contract 2019-19051600004), the IARPA HIATUS Program (contract 2022-22072200005), the NSF (Award 1928631), and an AI2 Young Investigator Award. Approved for Public Release, Distribution Unlimited. The views and conclusions contained herein are those of the authors and should not be interpreted as necessarily representing the official policies, either expressed or implied, of ODNI, DARPA, IARPA, NSF, or the U.S. Government.

{\small
\bibliographystyle{ieee_fullname}
\bibliography{egbib}
}
\clearpage
\appendix
\section{Dataset Statistics} \label{Sec: dataset stat}
Table \ref{tab: data stat} depicts detailed statistics for all datasets. For each dataset, we provide in parentheses a one-word description of the type of classes it contains, which we refer to as \textit{super class} of a dataset. We use the same train/dev/test splits of Food-101, Aircraft, Flower-102, UCF-101, and DTD provided by CoOp \cite{zhou2022learning}. For CUB, we randomly sample 10 training images for each category as the development set. For CIFAR-10 and CIFAR-100, we randomly split 10\% of the training data as the dev set. For HAM10000, we adopt 80/10/10 splits on the images of each class. For ImageNet, we only evaluate the dev set.
\begin{table}[!ht]
\centering
\resizebox{8.3cm}{!}{%
\begin{tabular}{lcccc}
\Xhline{3\arrayrulewidth}  
\multirow{2}{*}{\textbf{Name}} & \multirow{2}{*}{\begin{tabular}[c]{@{}c@{}}\textbf{n. of}\\\textbf{class}\end{tabular}} & \multicolumn{3}{c}{\textbf{n. of Images}} \\\cline{3-5}
                &        & Train      & Dev      & Test     \\\hline
Food-101 (food)           & 101   & 50,500 & 20,200 & 30,300 \\
FGVC-Aircraft (aircraft)  & 102   & 3,334 & 3,333 & 3,333 \\
Flower-102 (flower)       & 102   & 4,093 & 1,633 & 2,463 \\
CUB-200-2011 (bird)       & 200   & 3,994 & 2,000 & 5,794 \\
UCF-101 (action)          & 101   & 7,639 & 1,898 & 3,783\\
DTD (texture)             & 47    & 2,820 & 1,128 & 1,692 \\
HAM10000 (lesion)         & 7     & 8,010 & 1,000 & 1,005 \\
RESISC45 (scene)          & 45    & 3,150 & 3,150 & 25,200 \\
CIFAR-10 (object)         & 10    & 45,000 & 5,000 & 10,000 \\
CIFAR-100 (object)        & 100   & 45,000 & 5,000 & 10,000 \\
ImageNet (object)         & 1,000  &  1,281,167  &  50,000  &   -      \\\Xhline{3\arrayrulewidth}  
\end{tabular}
}
\caption{Detailed statistics of the 11 datasets. The text in parentheses that follows the dataset name corresponds to the super class name, which is used to remove class names in concepts.}
\label{tab: data stat}
\vspace{-0.5cm}
\end{table}

\section{Implementation Details} \label{appendix: details}
\subsection{Linear Probe}
Following CLIP's implementation of Linear Probe, we use the encoded images, before their projection to the vision-text embedding space, as input to the classifier. We use sklearn's L-BFGS implementation of logistic regression with 1,000 maximum iterations. To determine the best performing values for the L2 regularization strength $C$, we perform binary search on the validation set initialized with $[1e^6, 1e^4, 1e^2, 1, 1e^{-2}, 1e^{-4}, 1e^{-6}]$. After determining the left and right bounds of $C$, we iteratively halve the interval with 8 steps to get the final hyperparameter value. We compare our Linear Probe results on ImageNet with CoOp. To perform a fair comparison, we select CLIP-RN50 as the vision encoder and perform 3 random runs to select the few shot images. As shown in Table \ref{tab: linear probe}, we marginally outperform CoOp in all data settings. 

\begin{table}[!ht]
\centering
\small
\begin{tabular}{cccccc}
\Xhline{3\arrayrulewidth}  
\textbf{\# of shots} & \textbf{1} & \textbf{2} & \textbf{4} & \textbf{8} & \textbf{16}  \\ \hline
CoOp     & 22.07 & 31.95 & 41.29 & 49.55 & 55.87     \\
Ours    &  \textbf{\textbf{22.26}} & \textbf{32.28} & \textbf{41.57} & \textbf{49.80} & \textbf{55.92}  \\\Xhline{3\arrayrulewidth}  
\end{tabular}

\caption{Compare linear probe performance on ImageNet with CoOp. All experiments are based on CLIP-RN50, and we report the average score of 3 random runs.}
\label{tab: linear probe}
\vspace{-0.5cm}
\end{table}

\subsection{Prompt}
\label{app:prompt}
Table \ref{table: prompts} presents the prompts used to query GPT-3. We design 5 general prompts and 5 additional prompts for UCF-101. The general prompts are used for all datasets, with a slight modification: we add the super-class name that describes the type of data present in more fine-grained datasets. For example, when prompting for Flower-102, we add the super class name \textit{flower} after each class name. In this way we reduce ambiguity problems: e.g., for the class \textit{bishop of llandaff}, without the super class name, GPT-3 returns results for \textit{bishop} instead of the  \textit{flower}. While this approach reduces ambiguities, it does not completely eliminate them. 
For example, we found that GPT-3 generates sentences about the \textit{mouse} (device), but in fact, the class \textit{mouse} on ImageNet refers to the animal. Future work can explore better prompting methods, such as providing a detailed definition for each class or designing customized prompts for each dataset.
\begin{table}[!ht]
\centering
\small
\begin{tabular}{l}
\Xhline{3\arrayrulewidth}
\multicolumn{1}{c}{\textbf{General Prompt Template}}  \\ \hline
1. describe what the \texttt{[CLASS NAME]} looks like: \\
2. describe the appearance of the \texttt{[CLASS NAME]}: \\
3. describe the color of the \texttt{[CLASS NAME]}: \\
4. describe the pattern of the \texttt{[CLASS NAME]}: \\
5. describe the shape of the \texttt{[CLASS NAME]}: \\\hline \hline
\multicolumn{1}{c}{\textbf{UCF-101 Prompt Template}}  \\ \hline

1. describe what the \texttt{[CLASS NAME]} looks like: \\
2. describe the appearance of the \texttt{[CLASS NAME]}: \\
3. describe how to perform the \texttt{[CLASS NAME]}: \\
4. describe a person performing the \texttt{[CLASS NAME]}: \\
5. describe what can you see when a person is \\ ~~~~~performing the \texttt{[CLASS NAME]}: \\
\Xhline{3\arrayrulewidth}
\end{tabular}
\caption{The prompt templates used to generate the raw sentences from GPT-3. The UCF-101 has a different set of prompts, while the other datasets share the same set of general templates.}
\label{table: prompts}
\vspace{-0.4cm}
\end{table}
\begin{table*}[!t]
\centering
\resizebox{17cm}{!}{%
\begin{tabular}{cc|cccccc|cccccc}
\Xhline{3\arrayrulewidth}
\multirow{2}{*}{\textbf{Dataset}}  & \multirow{2}{*}{\textbf{Method}} & \multicolumn{6}{c|}{\textbf{Dev}}   & \multicolumn{6}{c}{\textbf{Test}}  \\ \cline{3-14}
& & \textbf{1} & \textbf{2} & \textbf{4} & \textbf{8} & \textbf{16} & \textbf{Full} & \textbf{1} & \textbf{2} & \textbf{4} & \textbf{8} & \textbf{16} & \textbf{Full} \\\hline
\multirow{2}{*}{Food-101} & Linear Prob  
& 58.04 & 75.24 & 84.16 & \textbf{87.48} & \textbf{89.87} & \textbf{93.11} & 57.75 & 75.34 & 84.21 & \textbf{87.90} & \textbf{90.02} & \textbf{93.17}      \\
                          & LaBo (Ours)             
& \textbf{80.32} & \textbf{84.15} & \textbf{85.76} & 87.07 & 88.74 & 92.53 & \textbf{80.41} & \textbf{84.05} & \textbf{85.68} & 87.39 & 88.77 & 92.45     \\\hline

\multirow{2}{*}{Aircraft} & Linear Prob  
& 27.63 & 34.86 & 41.40 & \textbf{49.72} & \textbf{57.91} & \textbf{62.89} & 28.26 & 35.07 & \textbf{41.55} & \textbf{50.26} & \textbf{56.38} & \textbf{64.03}      \\
                          & LaBo (Ours)             
& \textbf{33.12} & \textbf{35.97} & \textbf{42.90} & 49.08 & 56.41 & 61.96 & \textbf{32.73} & \textbf{37.71} & 41.04 & 48.81 & 54.97 & 61.42     \\\hline

\multirow{2}{*}{Flower-102} & Linear Prob  
& \textbf{89.20} & \textbf{94.06} & \textbf{97.00} & \textbf{98.40} & \textbf{98.91} & \textbf{99.11} & \textbf{88.06} & \textbf{93.65} & \textbf{97.67} & \textbf{98.56} & \textbf{99.32} & \textbf{99.45}      \\
                          & LaBo (Ours)             
& 82.24 & 88.18 & 94.92 & 96.20 & 98.16 & 98.65 & 82.05 & 90.09 & 95.21 & 97.08 & 98.66 & 99.35    \\\hline

\multirow{2}{*}{CUB} & Linear Prob  
& 48.55 & 60.40 & \textbf{72.50} & \textbf{78.25} & \textbf{83.35} & \textbf{83.60} & 47.69 & 61.06 & \textbf{72.82} & \textbf{79.60} & \textbf{83.74} & \textbf{84.54}      \\
                          & LaBo (Ours)             
& \textbf{55.20} & \textbf{64.80} & 72.45 & 76.55 & 79.90 & 81.00 & \textbf{54.19} & \textbf{64.60} & 71.21 & 77.22 & 80.69 & 81.90     \\\hline

\multirow{2}{*}{UCF-101} & Linear Prob  
& 65.54 & 76.34 & 85.83 & 90.25 & \textbf{93.63} & \textbf{98.63} & 60.56 & 73.22 & 80.62 & 85.70 & \textbf{87.63} & \textbf{90.67}      \\
                          & LaBo (Ours)             
& \textbf{80.72} & \textbf{83.77} & \textbf{88.46} & \textbf{90.73} & 93.05 & 97.68 & \textbf{78.75} & \textbf{82.05} & \textbf{84.56} & \textbf{86.39} & 87.39 & 90.11     \\\hline

\multirow{2}{*}{DTD} & Linear Prob 
& 43.62 & 53.19 & 60.55 & \textbf{68.79} & \textbf{74.47} & \textbf{80.50} & 41.67 & 51.71 & 60.76 & \textbf{69.03} & \textbf{74.70} & \textbf{81.68}      \\
                          & LaBo (Ours)             
& \textbf{55.59} & \textbf{56.47} & \textbf{62.15} & 68.44 & 70.92 & 76.86 & \textbf{53.61} & \textbf{55.26} & \textbf{61.17} & 66.43 & 70.21 & 77.30     \\\hline

\multirow{2}{*}{HAM10000} & Linear Prob  
& 32.30 & \textbf{55.40} & 45.40 & 50.90 & \textbf{63.10} & \textbf{84.40} & 33.13 & \textbf{55.32} & 44.48 & 48.26 & \textbf{61.69} & \textbf{83.18}     \\
                          & LaBo (Ours)             
& \textbf{34.90} & 46.40 & \textbf{45.80} & \textbf{54.40} & 58.20 & 81.40 & \textbf{36.62} & 45.17 & \textbf{45.87} & \textbf{52.04} & 55.72 & 81.39     \\\hline

\multirow{2}{*}{RESISC45} & Linear Prob  
& 68.62 & \textbf{79.10} & \textbf{86.72} & \textbf{89.89} & \textbf{92.49} & \textbf{95.24} & 67.57 & \textbf{77.75} & \textbf{86.50} & \textbf{89.27} & \textbf{92.17} & \textbf{94.98}      \\
                          & LaBo (Ours)             
& \textbf{73.02} & 76.03 & 81.37 & 85.05 & 88.86 & 91.65 & \textbf{73.66} & 76.11 & 81.40 & 85.71 & 88.63 & 91.22     \\\hline

\multirow{2}{*}{CIFAR-10} & Linear Prob  
& 62.36 & 80.32 & 92.94 & \textbf{95.36} & \textbf{96.06} & \textbf{98.16} & 62.44 & 80.27 & 92.54 & \textbf{95.14} & \textbf{95.90} & \textbf{98.10}      \\
                          & LaBo (Ours)             
& \textbf{91.24} & \textbf{91.04} & \textbf{92.98} & 94.40 & 95.06 & 97.90 & \textbf{91.06} & \textbf{90.79} & \textbf{93.03} & 94.11 & 94.93 & 97.75     \\\hline

\multirow{2}{*}{CIFAR-100} & Linear Prob  
& 39.66 & 57.84 & 70.06 & \textbf{76.52} & \textbf{80.34} & \textbf{87.70} & 39.26 & 57.35 & 69.73 & \textbf{76.22} & \textbf{80.16} & \textbf{87.48}      \\
                          & LaBo (Ours)             
& \textbf{62.84} & \textbf{66.56} & \textbf{71.78} & 75.30 & 78.08 & 86.82 & \textbf{62.73} & \textbf{65.80} & \textbf{70.82} & 74.49 & 77.67 & 86.04     \\\hline

\multirow{2}{*}{ImageNet} & Linear Prob  
& 42.25 & 55.71 & \textbf{64.80} & \textbf{71.23} & \textbf{75.08} & 83.90   & -  & -  & -  & -  &  -  &   -   \\
                          & LaBo (Ours)             
&  \textbf{51.09} & \textbf{57.43} & 62.94 & 68.45 & 72.60 & \textbf{83.97}   & -  & -  & -  & -  &  -  &   -    \\\hline

\multirow{2}{*}{Average} & Linear Prob  
& 52.53 & 65.68 & 72.85 & \textbf{77.89} & \textbf{82.29} & \textbf{87.93} & 51.69 & 65.13 & \textbf{72.33} & \textbf{77.38} & \textbf{81.53} & \textbf{87.38}      \\
                          & LaBo (Ours)             
& \textbf{63.66} & \textbf{68.25} & \textbf{72.86} & 76.88 & 80.00 & 86.40 & \textbf{63.35} & \textbf{68.10} & 72.08 & 76.19 & 79.11 & 85.72     \\ \Xhline{3\arrayrulewidth}
\end{tabular}
}
\caption{Full results of Linear Prob and LaBo on the development and test sets of 11 datasets.}
\label{tab: full results}
\vspace{-0.5cm}
\end{table*}

\subsection{T5 concept extractor}
The raw outputs of language models are long sentences and sometimes contain class names that need to be removed from the bottlenecks for the sake of interpretability. 
For example, GPT-3 generates a sentence ``\textit{The hen is brown and has a white chest.}'' for the class \textit{hen}, which could be decomposed to two concepts: ``\textit{brown}'' and ``\textit{white chest}''.
We annotate a random sample of 100  sentence-concepts pairs from each of the following datasets: Food-101, CIFAR-100, Aircraft, Flower, and ImageNet. In total, we collect 500 sentences. An example annotation is depicted below:
\vspace{-0.1cm}
\begin{center}
\begin{tcolorbox} [top=2pt,bottom=2pt, width=\linewidth, boxrule=1pt]
{\small {\fontfamily{put}\selectfont
The 737-400 has a long and slender fuselage 
with tapered wings and a small tail. (737-400)

long and slender fuselage; tapered wings; small tail
}
\par}
\end{tcolorbox}
\end{center}
\vspace{-0.1cm}
The class name is concatenated with the raw sentence, and the concepts are separated by semicolons. We train a T5-large model \cite{2020t5} using the \href{https://huggingface.co/docs/transformers/model_doc/t5}{Huggingface API}. We add a task prefix - ``\textit{extract concepts from sentence: }'' for each example. We train the model with Adam optimizer for 5 epochs, setting the batch size to 8 and learning rate to $1e^{-5}$.

\subsection{Remove Class Name}
After extracting the short concepts using T5, some still contain class names. To ensure there are no class names in the bottleneck, we design two heuristics: (1) If we find the class name in the concept using string match, we replace it with the super class name\footnote{The super class name depends on the datasets. For example, the super class name for the Flower-102 dataset is \textit{flower} (see Table \ref{tab: data stat}).}, e.g., the concept ``\textit{leaves of the \textbf{orange dahlia} are long and narrow}'' for the class \textit{orange dahlia} in Flower-102 is modified as ``\textit{leaves of the \textbf{flower} are long and narrow}''. (2) For class names with multiple tokens, the tokens are not always in the same order as the class name. In this case, if a concept with all tokens for the class name is present, we remove it. For instance, the concept ``\textit{a \textbf{cake} made of \textbf{carrot}}'' for the class \textit{carrot cake} will be deleted. The two heuristics are applied to each concept by considering all class names in the dataset. 

\subsection{Hyperparameters}
\label{app:hyperparameters}
We apply grid search with 5 runs to find the best weights for the submodular function for different datasets and shots. We determine the learning rate and batch size by monitoring the validation accuracy with \href{https://wandb.ai}{wandb}. Table \ref{tab: hyperparameters} lists all the hyperparameters of our best-performing models. 

\subsection{Other Details}
\noindent \textbf{GPT-3 Generation.} Generating 500 sentences for one class takes around 5 minutes by calling the OpenAI APIs. The price of GPT-3-Davinci is \$ 0.02 / 1k tokens, and it costs about \$ 0.2 for each class.
\medbreak
\noindent \textbf{Running Time.} Because we use CLIP with frozen weights, we only need to extract the image features once and reuse them in the rest experiments. Since we only fit a single linear layer, our training time is low. For example, training the full ImageNet for one epoch on an NVIDIA RTX A6000 takes less than 1 minute.
\medbreak
\noindent \textbf{Full Results.} The full numerical results are shown in Table \ref{tab: full results}. Both validation and test accuracy are provided.

\begin{figure*}[!t]
  \begin{subfigure}{0.33\textwidth}
    \includegraphics[width=.99\linewidth]{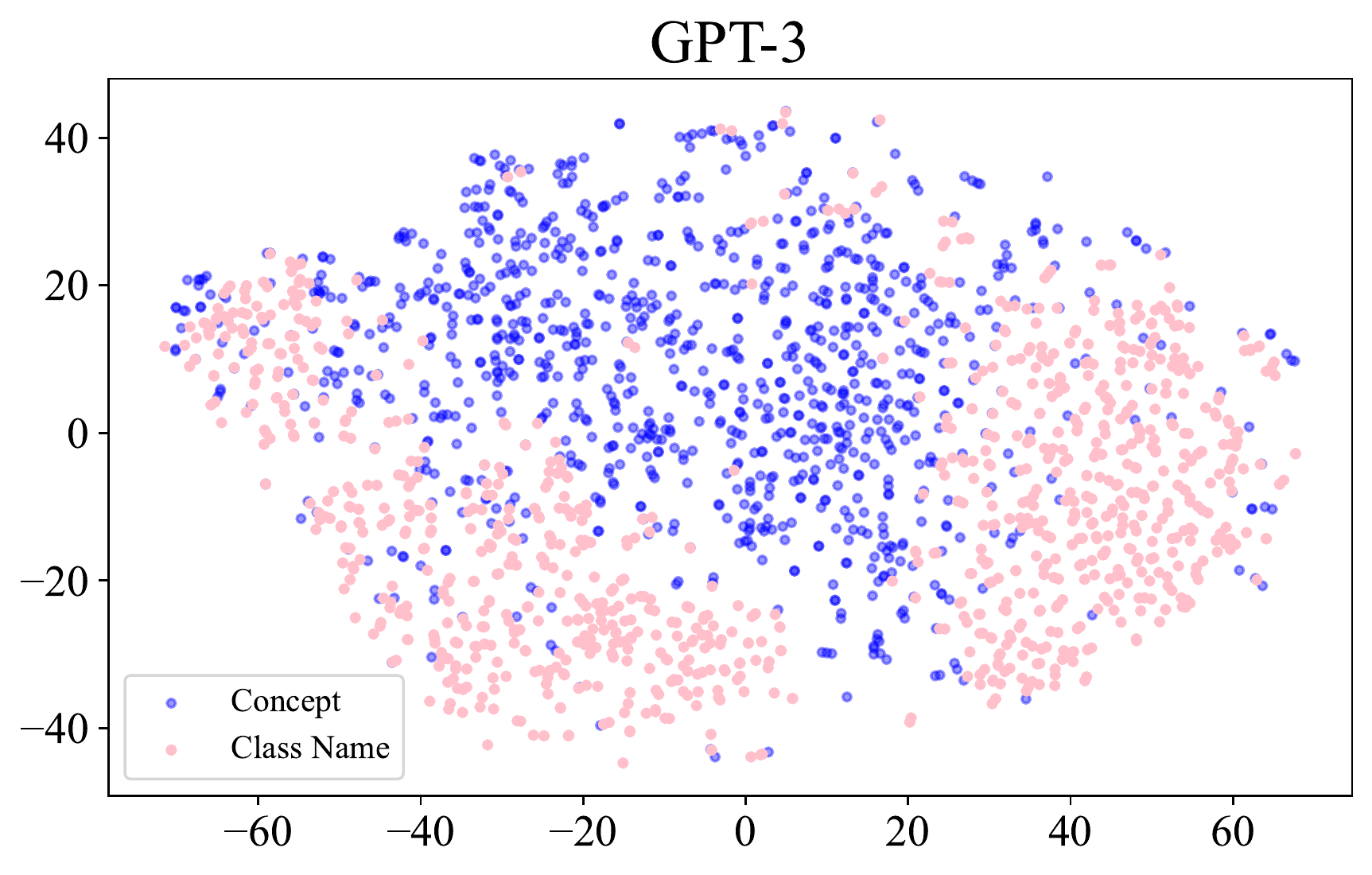}
  \end{subfigure}
  \begin{subfigure}{0.33\textwidth}
    \includegraphics[width=.99\linewidth]{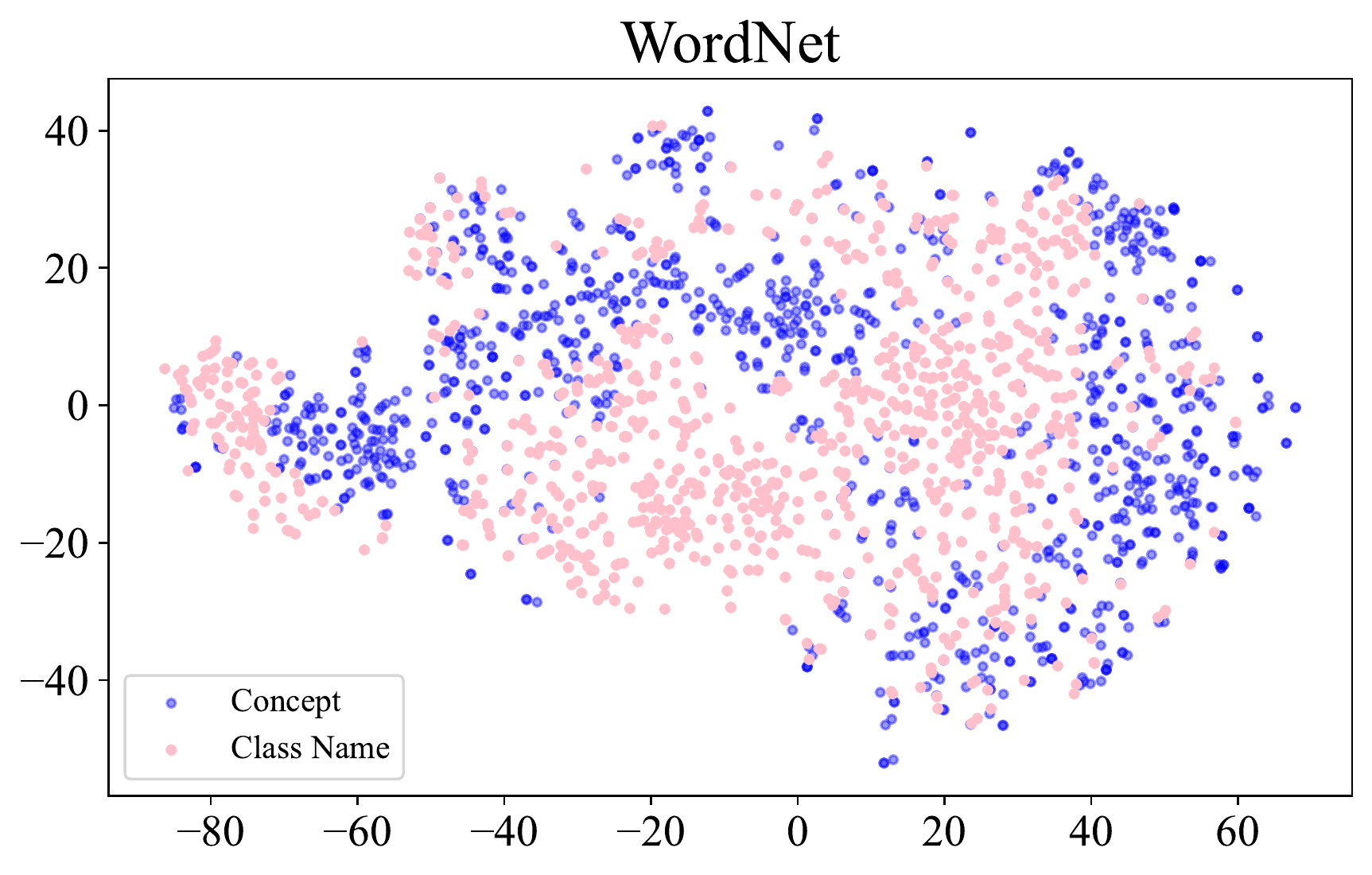}
  \end{subfigure}
  \begin{subfigure}{0.33\textwidth}
    \includegraphics[width=.99\linewidth]{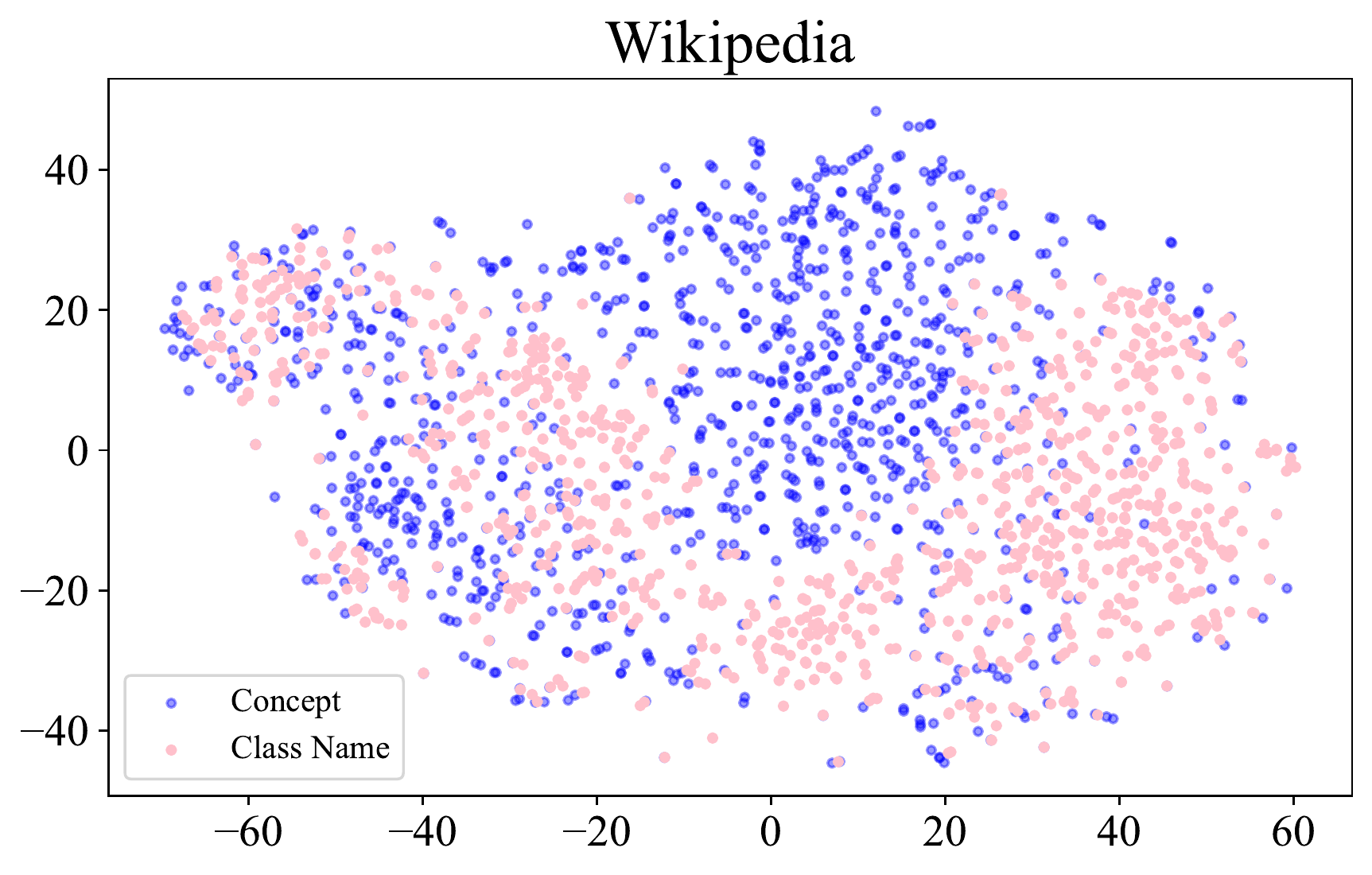}
  \end{subfigure}
  \caption{t-SNE visualization of the embeddings of concepts (blue) and class names (pink) on ImageNet. For the three bottlenecks constructed from GPT-3, WordNet, and Wikipedia, we visualize the top-1 concept of each class ranked by the weights of the linear function.}
  \label{fig: tsne}
  \vspace{-0.3cm}
\end{figure*}

\section{Additional Analysis}
\subsection{Activation Function}
We ablate the impact of the softmax activation by removing it or replacing it with other activation functions such as ReLU and sigmoid. As shown in Table \ref{tab: activation function}, not using an activation function significantly hurts performance, while using other activation functions performs poorly compared to softmax.
\begin{table}[!t]
\centering
\resizebox{8.3cm}{!}{%
\begin{tabular}{ccccccc}
\Xhline{3\arrayrulewidth}  
\textbf{Activation} & \textbf{1} & \textbf{2} & \textbf{4} & \textbf{8} & \textbf{16} & \textbf{Full} \\ \hline
-        & 52.66 & 58.01 & 63.02 & 68.93 & 73.52 & 81.32  \\
\texttt{relu}    & 50.40 & 53.53 & 56.61 & 59.82 & 61.75 & 68.01  \\
\texttt{sigmoid} & 52.15 & 57.86 & 62.59 & 69.08 & 73.43 & 81.42  \\
\texttt{softmax} & \textbf{63.03} & \textbf{67.79} & \textbf{71.88} & \textbf{76.08} & \textbf{79.10} & \textbf{85.71} \\\Xhline{3\arrayrulewidth}  
\end{tabular}
}
\caption{Compare different activation functions. We report the mean accuracy across the 11 datasets.}
\label{tab: activation function}
\vspace{-0.1cm}
\end{table}

\subsection{Language Model Size vs. Performace}
We experiment with different sizes of GPT-3: Curie, Babbage, and Ada (sorted from larger to smaller). Figure \ref{tab: lm size} compares the different GPT-3 variants on ImageNet, showing that larger language models result in better performance, especially in the few show settings. However, there is only a marginal difference in performance when enough data is available. 
\begin{table}[!t]
\centering
\resizebox{8.3cm}{!}{%
\begin{tabular}{lcccccc}
\Xhline{3\arrayrulewidth}  
\textbf{GPT-3 type} & \textbf{1} & \textbf{2} & \textbf{4} & \textbf{8} & \textbf{16} & \textbf{Full} \\ \hline
Davinci (175B)         & \textbf{51.09}	& \textbf{57.43}	& \textbf{62.94}	& \textbf{68.45}	& \textbf{72.60}	& 83.97     \\
Curie (13B)   & 45.75 & 53.89 & 60.36 & 66.96 & 71.65 & \textbf{84.00} \\
Babbage (6.7B)   & 44.61 & 52.91 & 60.22 & 67.06 & 71.66 & 83.86 \\
Ada (2.7B)   & 43.12 & 53.26 & 60.99 & 67.90 & 72.42 & 83.96 \\\Xhline{3\arrayrulewidth}  
\end{tabular}
}
\caption{The performance of LaBo on ImageNet using different sizes of GPT-3 to generate concepts. The number in the parenthesis is the number of parameters of the corresponding language model.}
\label{tab: lm size}
\vspace{-0.3cm}
\end{table}

\subsection{Performance of Human-Written Text}
Table \ref{tab: human text} compares the performance of LaBo between using GPT-3 generated concepts and human-designed concepts sourced from WordNet and Wikipedia. We observe that GPT-3 generated concepts outperform human-written ones in 1-shot experiments, while there is less than 1\% drop in performance on average in larger data settings. In addition, our human evaluation on Imagenet (see Figure \ref{fig: imagenet human eval} and \ref{fig: invalid concepts} in Section \ref{sec:human_eval}) shows that humans judge the quality of GPT-3 generated concepts to be better than that of human-designed. 

 We visualize the embeddings of concepts and class names using t-SNE \cite{van2008visualizing} to identify the reason behind the perceived higher quality of GPT-3 concepts. We encode the 1,000 class names of ImageNet using the CLIP text encoder along with the top-1 concept of each class (1,000 concepts in total) from each bottleneck (LaBo, WordNet, and Wikipedia). Figure \ref{fig: tsne} reflects that compared to GPT-3, the embeddings of WordNet and Wikipedia concepts have a higher overlap with the embeddings of class names. In other words, Wikipedia and WordNet concepts are more likely to replicate the text features of class names rather than describe the class. This explains why human-written text has higher accuracy but is less interpretable.
\begin{table}[!t]
\centering
\resizebox{8.3cm}{!}{%
\begin{tabular}{lcccccc}
\Xhline{3\arrayrulewidth}  
\textbf{Concept Source} & \textbf{1} & \textbf{2} & \textbf{4} & \textbf{8} & \textbf{16} & \textbf{Full} \\ \hline
GPT-3       & \textbf{51.09} & 57.43 & 62.94 & 68.45 & 72.60 & 83.97   \\
Wikipedia   & 48.76 & 56.73 & 63.00 & 68.96 & 73.07 & \textbf{84.07} \\
WordNet     & 49.37 & \textbf{57.84} & \textbf{64.10} & \textbf{69.92} & \textbf{73.35} & 83.93 \\\Xhline{3\arrayrulewidth}  
\end{tabular}
}
\caption{The performance of LaBo on ImageNet using different sources of concepts to construct the bottlenecks.}
\label{tab: human text}
\vspace{-0.1cm}
\end{table}

\begin{table}[!t]
\centering
\resizebox{8.3cm}{!}{%
\begin{tabular}{ccccccc}
\Xhline{3\arrayrulewidth}  
\textbf{Method} & \textbf{w/ cls} & \textbf{Aircraft} & \textbf{Food} & \textbf{Flower} & \textbf{DTD} & \textbf{UCF} \\ \hline
LP         &   -            &    39.42 & 76.99 & 95.89 & 68.74 & 80.04       \\ \hline
LaBo       &   \xmark       &  37.29 & 76.04 & 92.37 & 64.78 & 80.07       \\
CoOp \cite{zhou2022learning}      &   \cmark       &  33.22 & \textbf{78.45} & \textbf{94.97} & \textbf{65.37} & 78.66     \\
LaBo$^\dagger$       &   \cmark       &  \textbf{37.53} & 77.83 & 93.18  &  \textbf{65.37} & \textbf{80.10}\\\Xhline{3\arrayrulewidth}  
\end{tabular}
}
\caption{Compare LaBo with prompt tuning methods on 5 datasets (16 shots). w/ cls stands for using class names in the context. LaBo$^\dagger$ is our method without removing the class names in the concepts. All methods use CLIP-ViT-B/32 as the vision backbone.}
\label{tab: coop}
\vspace{-.3cm}
\end{table}
\subsection{Comparison with the Prompt Tuning Method}
Table \ref{tab: coop} compares the performance between LaBo and CoOp\cite{zhou2022learning}, which employs a soft prompt tuning method (not interpretable) on five datasets. Even though LaBo does not use class names, its performance is similar to that of CoOp. Adding class names to LaBo leads to performance gains, such that it outperforms CoOp on Aircraft and UCF-101. 
\begin{figure*}[!t]
\centering
    \includegraphics[width=17.3cm]{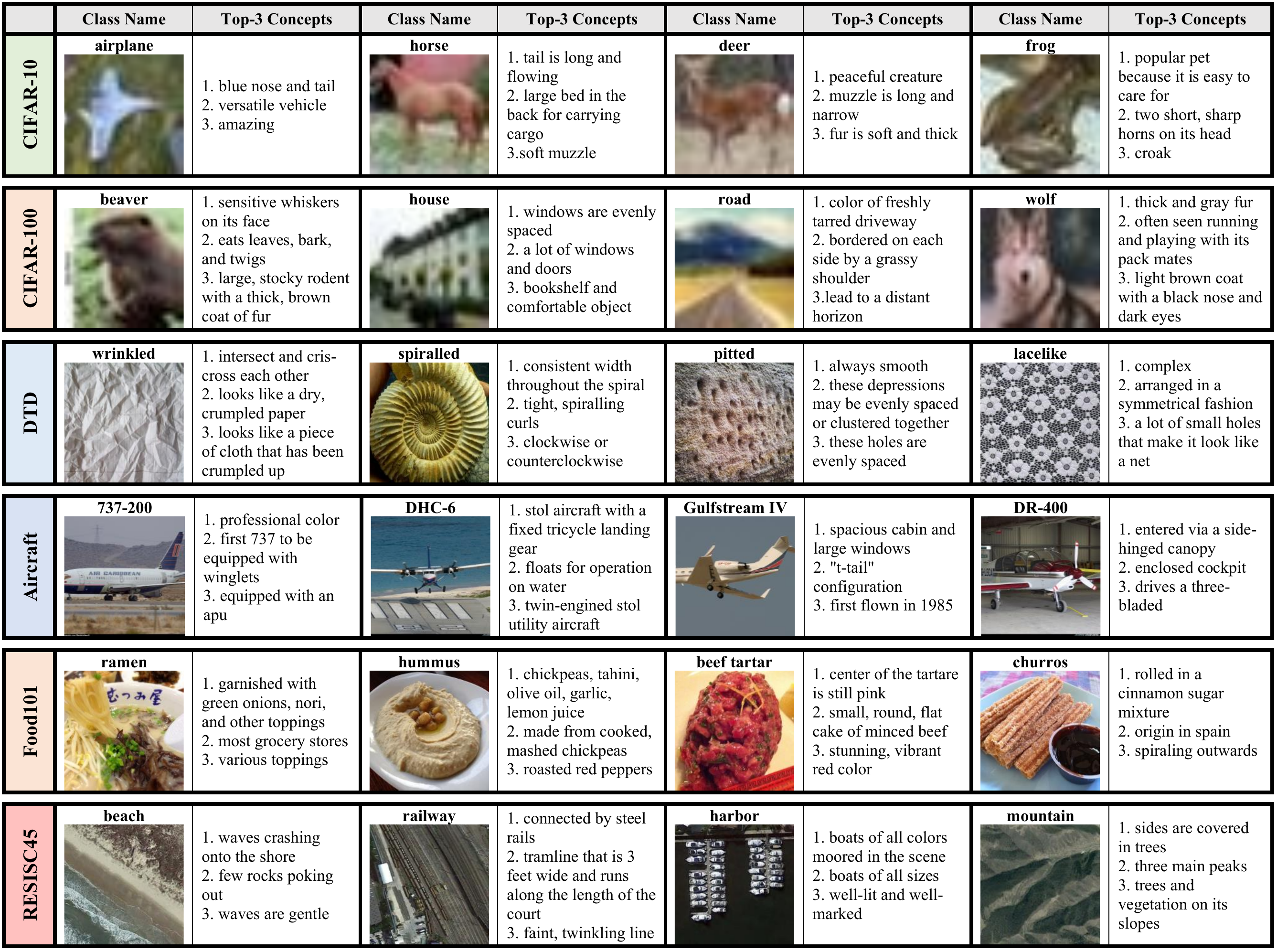}
    \caption{Additional qualitative examples for CIFAR-10, CIFAR-100, DTD, Aircraft, Food101 and RESISC45.}
    \label{fig: qual examples 2}
\end{figure*}

\section{Human Evaluation }
\label{app:human_eval}
\begin{table*}[!ht]
\centering
\resizebox{16.5cm}{!}{%
\begin{tabular}{lcllcllcllcllcllcllcllcllcllcll}
\Xhline{3\arrayrulewidth}  
\multicolumn{1}{c|}{} & \multicolumn{3}{c}{\textbf{Food}} & \multicolumn{3}{c}{\textbf{Aircraft}} & \multicolumn{3}{c}{\textbf{HAM10K}} & \multicolumn{3}{c}{\textbf{RESISC}} & \multicolumn{3}{c}{\textbf{Flower}} & \multicolumn{3}{c}{\textbf{CUB}} & \multicolumn{3}{c}{\textbf{UCF}} & \multicolumn{3}{c}{\textbf{DTD}} & \multicolumn{3}{c}{\textbf{CIFAR10}} & \multicolumn{3}{c}{\textbf{CIFAR100}} \\ \hline
\multicolumn{1}{l|}{\textbf{Factuality} $\uparrow$} & \multicolumn{3}{c}{P@10} & \multicolumn{3}{c}{P@10} & \multicolumn{3}{c}{P@10} & \multicolumn{3}{c}{P@10} & \multicolumn{3}{c}{P@8} & \multicolumn{3}{c}{P@10} & \multicolumn{3}{c}{P@10} & \multicolumn{3}{c}{P@10} & \multicolumn{3}{c}{P@10} & \multicolumn{3}{c}{P@10} \\ \cline{2-31} 
\multicolumn{1}{l|}{LaBo} & \multicolumn{3}{c}{\textbf{33.07}} & 
\multicolumn{3}{c}{\textbf{11.57}} & \multicolumn{3}{c}{15.05} & 
\multicolumn{3}{c}{14.80} & \multicolumn{3}{c}{11.48}
& \multicolumn{3}{c}{\textbf{27.97}} & \multicolumn{3}{c}{\textbf{37.78}} & \multicolumn{3}{c}{23.90} & \multicolumn{3}{c}{14.70} & \multicolumn{3}{c}{22.48} \\
\multicolumn{1}{l|}{w/o submod} & \multicolumn{3}{c}{27.08} & \multicolumn{3}{c}{8.10} & \multicolumn{3}{c}{9.57} & 
\multicolumn{3}{c}{\textbf{16.40}} & 
\multicolumn{3}{c}{\textbf{18.58}} & \multicolumn{3}{c}{23.12} & \multicolumn{3}{c}{37.22} & \multicolumn{3}{c}{\textbf{25.27}} & \multicolumn{3}{c}{\textbf{20.70}} & \multicolumn{3}{c}{\textbf{22.72}} \\
\multicolumn{1}{l|}{w/o LM} & \multicolumn{3}{c}{21.63} & \multicolumn{3}{c}{8.97} & \multicolumn{3}{c}{\textbf{19.71}} & \multicolumn{3}{c}{12.15} & \multicolumn{3}{c}{9.98} & \multicolumn{3}{c}{12.17} & \multicolumn{3}{c}{20.43} & \multicolumn{3}{c}{14.83} & \multicolumn{3}{c}{6.87} & \multicolumn{3}{c}{14.97} \\ \hline \hline
\multicolumn{1}{l|}{\textbf{Groundability}$\uparrow$} & \multicolumn{3}{c}{P@10} & \multicolumn{3}{c}{P@10} & \multicolumn{3}{c}{P@10} & \multicolumn{3}{c}{P@10} & \multicolumn{3}{c}{P@8} & \multicolumn{3}{c}{P@10} & \multicolumn{3}{c}{P@10} & \multicolumn{3}{c}{P@10} & \multicolumn{3}{c}{P@10} & \multicolumn{3}{c}{P@10} \\ \cline{2-31} 
\multicolumn{1}{l|}{LaBo} & \multicolumn{3}{c}{10.98} & \multicolumn{3}{c}{8.48} & \multicolumn{3}{c}{18.83} & \multicolumn{3}{c}{13.87} & \multicolumn{3}{c}{9.53} & \multicolumn{3}{c}{15.63} & \multicolumn{3}{c}{8.08} & \multicolumn{3}{c}{8.90} & \multicolumn{3}{c}{5.70} & \multicolumn{3}{c}{19.83} \\
\multicolumn{1}{l|}{w/o submod} & \multicolumn{3}{c}{\textbf{21.52}} & \multicolumn{3}{c}{\textbf{13.67}} & \multicolumn{3}{c}{17.22} & \multicolumn{3}{c}{\textbf{17.90}} & \multicolumn{3}{c}{\textbf{21.52}} & \multicolumn{3}{c}{23.07} & \multicolumn{3}{c}{\textbf{29.93}} & \multicolumn{3}{c}{20.02} & \multicolumn{3}{c}{\textbf{23.10}} & \multicolumn{3}{c}{21.78} \\
\multicolumn{1}{l|}{w/o LM} & \multicolumn{3}{c}{20.58} & \multicolumn{3}{c}{12.00} & \multicolumn{3}{c}{\textbf{20.00}} & \multicolumn{3}{c}{14.38} & \multicolumn{3}{c}{17.93} & \multicolumn{3}{c}{\textbf{25.02}} & \multicolumn{3}{c}{27.96} & \multicolumn{3}{c}{\textbf{20.31}} & \multicolumn{3}{c}{7.15} & \multicolumn{3}{c}{\textbf{27.04}} \\ \Xhline{3\arrayrulewidth}  
\end{tabular}
}
\caption{Analytic Factuality and Groundability for all datasets except Imagenet (see Figure \ref{fig: imagenet human eval})}
\label{tab:human_eval_all}
\vspace{-0.3cm}
\end{table*}

\begin{figure*}[!ht]
    \includegraphics[width=17cm]{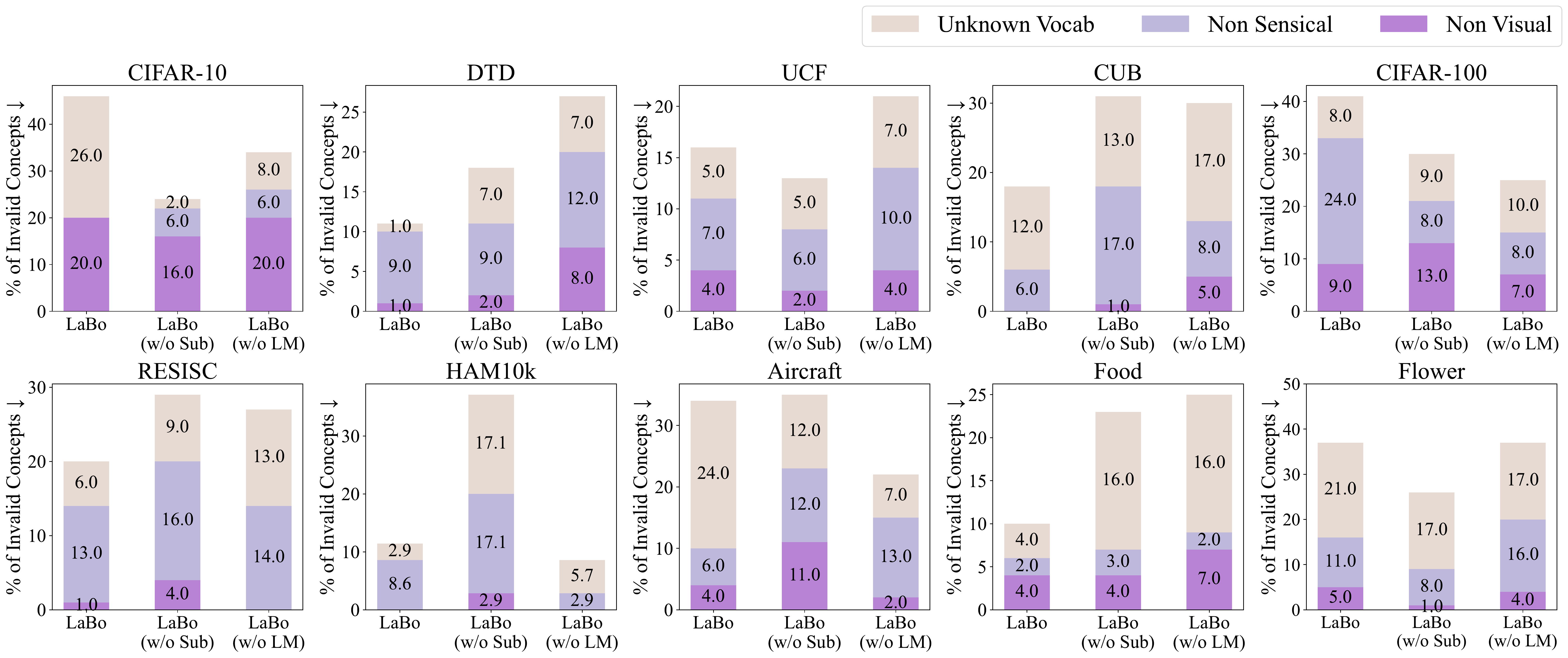}
    \caption{Percentage of invalid concepts identified by humans for different bottlenecks for all 10 datasets except ImageNet (see Figure \ref{fig: invalid concepts}). \textbf{Lower} percentage is better.}
    \label{fig: invalid concepts all}
    \vspace{-0.3cm}
\end{figure*}

\begin{figure}[!t]
\centering
    \includegraphics[width=\linewidth]{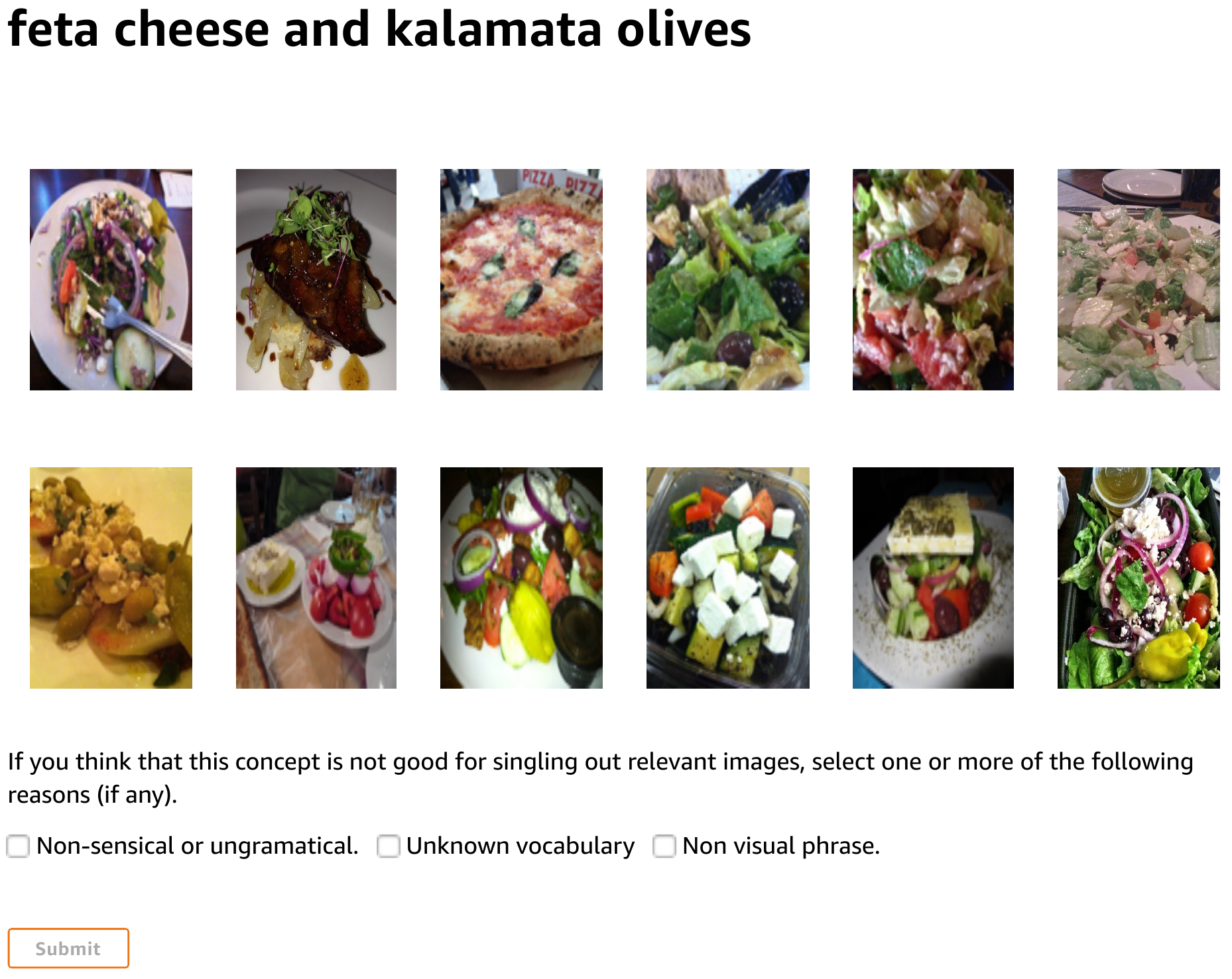}
    \caption{Sample user interface for measuring \textit{Factuality}. We provide 10 ground truth images with 2 control images randomly positioned. Annotators are required to select the images that can be described by the phrase. The user interface for \textit{Groundability} is identical, but the images presented are the top-10 images in the dataset sorted by CLIP\cite{radford2021learning} similarity score. }
    \label{fig: mturk_ui}
    \vspace{-0.3cm}
\end{figure}

\begin{figure*}[!t]
\centering
    \includegraphics[width=.9\linewidth]{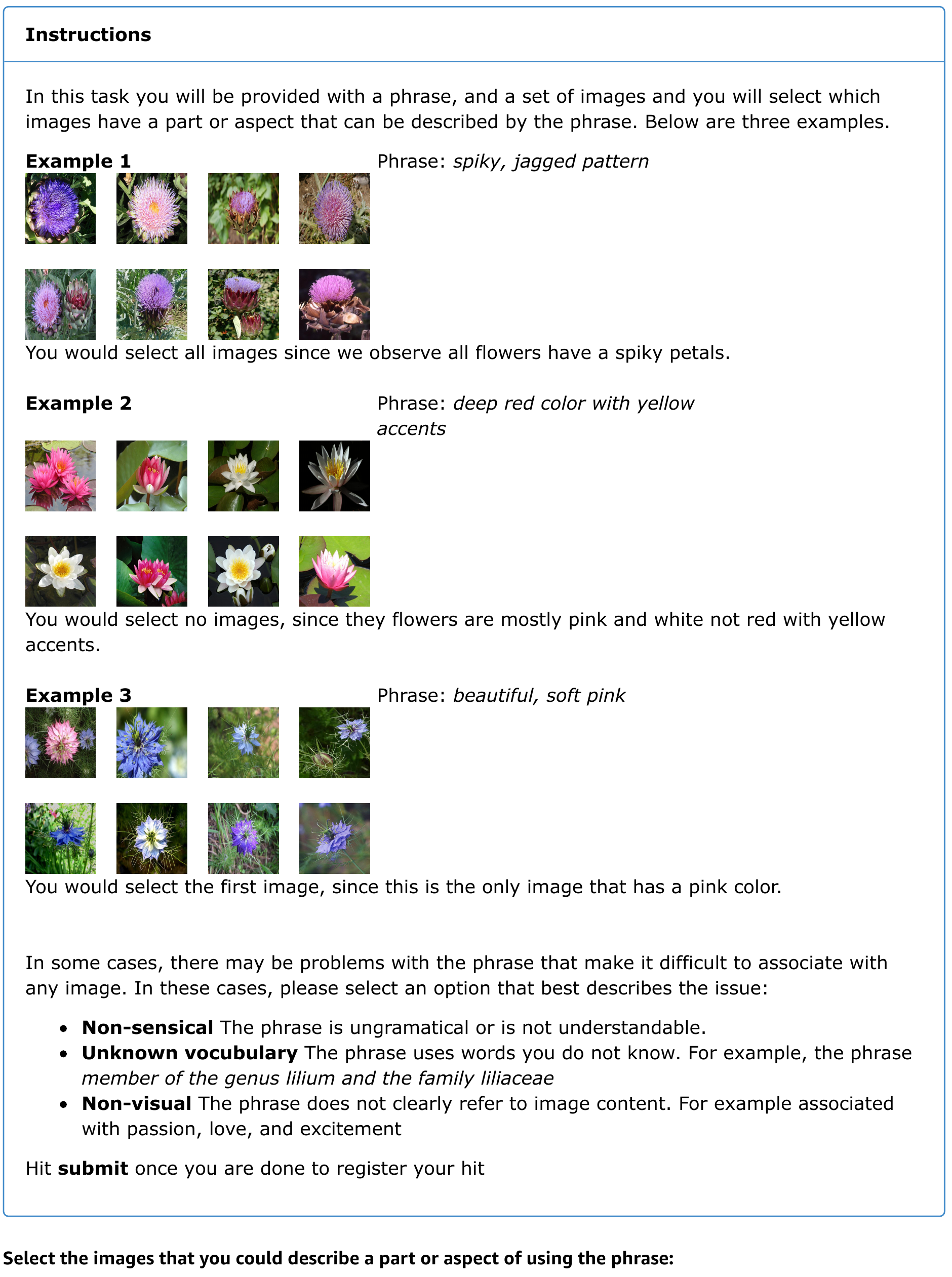}
    \caption{Instructions provided to annotators to compute \textit{Factuality} and \textit{Groundability}.}
    \label{fig: mturk_instructions}
\end{figure*}

We introduce two qualitative metrics to evaluate the automatically generated concept bottlenecks to highlight areas of possible improvement. We introduce two metrics that evaluate the bottleneck items along two dimensions: \textit{Factuality} and \textit{Groundability} (see Section \ref{sec:human_eval}). 
\medbreak
\noindent \textbf{Annotator Statistics.} Both metrics rely on human annotations, which we collect on \href{https://www.mturk.com}{Amazon Mechanical Turk}. To ensure confidence in the results, we collect 3 annotations per concept. Annotators are paid on average \$14.5 per hour, and the total cost of the annotation was \$2,100. Our rate was computed by estimating the time it takes to complete the task by 4 different control annotators.\footnote{Our focus group was graduate students. Since this is not representative of the average population, we doubled the time estimate.} In total, our task was completed by a diverse set of 477 annotators. The average pairwise annotator agreement for all annotated data without any pre-processing is 69.83\%. 
\medbreak
\noindent \textbf{Interface.} Figure \ref{fig: mturk_ui} displays the annotation interface. Given a concept phrase, annotators are prompted to select from 12 images, 10 of which correspond to the ground truth target corresponding to the concept, and 2 control images randomly sampled from other classes. The user interface was accompanied by a set of instructions presented in Figure \ref{fig: mturk_instructions}.
\medbreak
\noindent \textbf{Invalid Annotations.} In reporting \textit{Factuality} and \textit{Groundability}, we disregard annotations that select any of the control images unless all annotators failed the control for a particular concept. In total, we disregard 18\% of annotations for this reason. In reporting invalid concepts (non-visual, non-sensical, or unknown vocabulary), we consider all annotations but consider a bottleneck invalid if at least 2 out of 3 annotators agree. 
\medbreak
\noindent \textbf{Analytic Results.} Table \ref{tab:human_eval_all} displays analytic results of \textit{Factuality} and \textit{Groundability} for all datasets. Figure \ref{fig: invalid concepts all} presents the invalid concept distribution for all datasets separately. It is worth noting the high percentage of non-visual concepts in CIFAR-10 and CIFAR-100 compared to other datasets. We hypothesize that this reflects the annotators' inability to see the images clearly due to the low resolution (see Figure \ref{fig: qual examples 2}) rather than the lack of visual content in the concept. For example, the concepts ``small and black" and ``blue nose and tail" were annotated as non-visual for CIFAR-10, and the concepts ``color of trees and grass" and ``two large pincers on its front legs" for CIFAR-100.

\section{Qualitative Examples}
Figure \ref{fig: qual examples 2} shows the additional qualitative examples for the rest 6 datasets (CIFAR-10, CIFAR-100, DTD, Aircraft, Food101, and RESISC45).

\begin{table*}[!ht]
\centering
\resizebox{12cm}{!}{%
\begin{tabular}{ccccccc}
\Xhline{3\arrayrulewidth}  
 & \textbf{n. of shots} & \textbf{Bottleneck Size} & \textbf{Discriminability} ($\alpha$) & \textbf{Coverage} ($\beta$) & \textbf{Learning Rate} & \textbf{Batch Size} \\ \hline
\multirow{6}{*}{\rotatebox[origin=c]{90}{Food-101}} 
                      & 1     &  5,050  & $1e^7$  &  0.5 &  $1e^{-5}$  &   16  \\
                      & 2     &  5,050  & $1e^7$  &  1   &  $1e^{-4}$  &   32 \\
                      & 4     &  5,050  & $1e^7$  &  1   &  $1e^{-4}$  &   64 \\
                      & 8     &  5,050  & $1e^7$  &  1   &  $1e^{-4}$  &   128  \\
                      & 16    &  5,050  & $1e^7$  &  1   &  $1e^{-4}$  &   256 \\
                      & Full  &  5,050  & $1e^7$  &  5   &  $1e^{-5}$  &   1024 \\\hline
\multirow{6}{*}{\rotatebox[origin=c]{90}{Aircraft}} 
                      & 1     &  5,100  & $1e^7$  &  0.5  &  $5e^{-5}$  &  16  \\
                      & 2     &  5,100  & $1e^7$  &  1    &  $5e^{-5}$  &  32  \\
                      & 4     &  5,100  & $1e^7$  &  0.1  &  $5e^{-5}$  &  64  \\
                      & 8     &  5,100  & $1e^7$  &  0    &  $5e^{-5}$  &  128  \\
                      & 16    &  5,100  & $1e^7$  &  1    &  $5e^{-5}$  &  256  \\
                      & Full  &  5,100  & $1e^7$  &  0.5  &  $5e^{-5}$  &  256  \\\hline
\multirow{6}{*}{\rotatebox[origin=c]{90}{Flower-102}} 
                      & 1     &  2,050  & $1e^7$  &  10  & $1e^{-5}$   & 16   \\
                      & 2     &  2,050  & $1e^7$  &  100 & $1e^{-5}$   & 32  \\
                      & 4     &  2,050  & $1e^7$  &  10  & $1e^{-5}$   & 64   \\
                      & 8     &  2,050  & $1e^7$  &  10  & $1e^{-5}$   & 128  \\
                      & 16    &  2,050  & $1e^7$  &  1   & $1e^{-5}$   & 256  \\
                      & Full  &  2,050  & $1e^7$  &  1   & $1e^{-5}$      & 256   \\\hline
\multirow{6}{*}{\rotatebox[origin=c]{90}{CUB}} 
                      & 1     & 2,000  &  $1e^7$ &  0   & $5e^{-5}$   &  32  \\
                      & 2     & 2,000  &  $1e^7$ &  0   & $5e^{-5}$   &  64  \\
                      & 4     & 2,000  &  $1e^7$ &  0.1 & $5e^{-5}$   &  128  \\
                      & 8     & 2,000  &  $1e^7$ &  0   & $5e^{-5}$   &  256  \\
                      & 16    & 2,000  &  $1e^7$ &  1   & $5e^{-5}$   &  512  \\
                      & Full  & 2,000  &  $1e^7$ &  0.1 & $5e^{-5}$   &  512  \\\hline
\multirow{6}{*}{\rotatebox[origin=c]{90}{UCF-101}} 
                      & 1     & 5,050  &  $1e^7$ &  1   & $1e^{-5}$   &  8  \\
                      & 2     & 5,050  &  $1e^7$ &  1   & $1e^{-5}$   &  16  \\
                      & 4     & 5,050  &  $1e^7$ &  100 & $1e^{-5}$   &  32  \\
                      & 8     & 5,050  &  $1e^7$ &  10  & $1e^{-5}$   &  64  \\
                      & 16    & 5,050  &  $1e^7$ &  100 & $1e^{-5}$   &  128  \\
                      & Full  & 5,050  &  $1e^7$ &  10  & $1e^{-5}$   &  256  \\\hline
\multirow{6}{*}{\rotatebox[origin=c]{90}{DTD}} 
                      & 1     & 2,350  &  $1e^7$ &  10 & $1e^{-5}$   &  8  \\
                      & 2     & 2,350  &  $1e^7$ &  10 & $1e^{-5}$   &  16  \\
                      & 4     & 2,350  &  $1e^7$ &  5  & $1e^{-5}$   &  32  \\
                      & 8     & 2,350  &  $1e^7$ &  1  & $1e^{-5}$   &  64  \\
                      & 16    & 2,350  &  $1e^7$ &  2.5 & $5e^{-5}$  &  256  \\
                      & Full  & 2,350  &  $1e^7$ &  7.5 & $1e^{-4}$  &  512  \\\hline
\multirow{6}{*}{\rotatebox[origin=c]{90}{HAM10000}} 
                      & 1     & 350  & $1e^7$  &  0.1 & $1e^{-3}$   &  4  \\
                      & 2     & 350  & $1e^7$  &  0.1 & $1e^{-3}$   &  4  \\
                      & 4     & 350  & $1e^7$  &  1   & $1e^{-4}$   &  8  \\
                      & 8     & 350  & $1e^7$  &  10  & $1e^{-3}$   &  8  \\
                      & 16    & 350  & $1e^7$  &  15  & $1e^{-3}$   &  16  \\
                      & Full  & 350  & $1e^7$  &  0.1 & $5e^{-4}$   &  256  \\\hline
\multirow{6}{*}{\rotatebox[origin=c]{90}{RESISC45}} 
                      & 1     & 2,250  & $1e^7$  & 5   & $5e^{-5}$   &  8  \\
                      & 2     & 2,250  & $1e^7$  & 5   & $5e^{-5}$   &  16  \\
                      & 4     & 2,250  & $1e^7$  & 10  & $5e^{-5}$   &  32  \\
                      & 8     & 2,250  & $1e^7$  & 15  & $5e^{-5}$   &  64  \\
                      & 16    & 2,250  & $1e^7$  & 15  & $5e^{-5}$   &  128  \\
                      & Full  & 2,250  & $1e^7$  & 15  & $5e^{-5}$   &  256  \\\hline
\multirow{6}{*}{\rotatebox[origin=c]{90}{CIFAR-10}} 
                      & 1     & 500  & $1e^7$  & 1  & $1e^{-4}$   &  2  \\
                      & 2     & 500  & $1e^7$  & 5  & $5e^{-4}$   &  4  \\
                      & 4     & 500  & $1e^7$  & 5  & $1e^{-4}$   &  8  \\
                      & 8     & 500  & $1e^7$  & 1  & $1e^{-4}$   &  16  \\
                      & 16    & 500  & $1e^7$  & 10 & $1e^{-4}$   &  32  \\
                      & Full  & 500  & $1e^7$  & 5  & $1e^{-4}$   &  512  \\\hline
\multirow{6}{*}{\rotatebox[origin=c]{90}{CIFAR-100}} 
                      & 1     & 5,000  & $1e^7$  & 7.5  &  $1e^{-5}$  & 16   \\
                      & 2     & 5,000  & $1e^7$  & 2.5  &  $1e^{-5}$  & 32   \\
                      & 4     & 5,000  & $1e^7$  & 7.5  &  $1e^{-5}$  & 64   \\
                      & 8     & 5,000  & $1e^7$  & 7.5  &  $1e^{-5}$  & 128   \\
                      & 16    & 5,000  & $1e^7$  & 5  &  $1e^{-5}$  & 256   \\
                      & Full  & 5,000  & $1e^7$  & 0  &  $1e^{-5}$  & 512   \\\hline
\multirow{6}{*}{\rotatebox[origin=c]{90}{ImageNet}} 
                      & 1     & 50,000  & $1e^8$  & 0  &  $1e^{-5}$   & 128   \\
                      & 2     & 50,000  & $1e^8$  & 0  &  $1e^{-5}$   & 256   \\
                      & 4     & 50,000  & $1e^8$  & 0  &  $1e^{-5}$   & 256  \\
                      & 8     & 50,000  & $1e^8$  & 0  &  $1e^{-5}$   & 512   \\
                      & 16    & 50,000  & $1e^8$  & 0  &  $1e^{-5}$   & 1024   \\
                      & Full  & 50,000  & $1e^8$  & 0  &  $1e^{-5}$   &  2048  \\ \Xhline{3\arrayrulewidth}  
\end{tabular}
}
\caption{All hyperparameters used for the main experiments which are tuned on the development set.}
\label{tab: hyperparameters}
\end{table*}

\end{document}